\documentclass[10pt,journal,compsoc]{IEEEtran}

\usepackage[pdftex]{graphicx}
\usepackage{amsmath}
\usepackage{amssymb}
\usepackage{ctable}
\usepackage{color,colortbl}
\usepackage[ruled,vlined,linesnumbered]{algorithm2e}
\usepackage[symbol]{footmisc}
\usepackage{multirow}

\definecolor{Gray}{gray}{0.9}
\definecolor{White}{gray}{1}

\usepackage{xcolor}

\usepackage[pagebackref=true,breaklinks=true,colorlinks,bookmarks=false]{hyperref}

\ifCLASSOPTIONcompsoc
  \usepackage[nocompress]{cite}
\else
  \usepackage{cite}
\fi
\begin{document}
%
\title{HybrIK-X: Hybrid Analytical-Neural Inverse Kinematics for Whole-body Mesh Recovery}
%
%
%
%

\author{Jiefeng Li,
        Siyuan Bian,
        Chao Xu,
        Zhicun Chen,
        Lixin Yang,
        and~Cewu Lu$^{\dagger}$,~\IEEEmembership{Member,~IEEE}
\IEEEcompsocitemizethanks{\IEEEcompsocthanksitem Jiefeng Li, Siyuan Bian, Zhicun Chen, Lixin Yang and Cewu Lu are with the Department of Electrical and Computer Engineering, Shanghai Jiao Tong University, Shanghai, 200240, China. E-mail: \{{ljf\_likit}, {biansiyuan}, {zhicun\_chen}, {siriusyang}, {lucewu}\}@{sjtu.edu.cn}.
\IEEEcompsocthanksitem Chao Xu is with Xiaoice Company, Shanghai, China.  E-mail: \{{xuchao.19962007}\}@{sjtu.edu.cn}.
\IEEEcompsocthanksitem Corresponding author: Cewu Lu.
}
}

%
%

\markboth{Journal of \LaTeX\ Class Files, ~Vol.~XX, No.~XX, XXX. XXXX}%
{****for peer review only****}
%



\IEEEtitleabstractindextext{%
\begin{abstract}
  Recovering whole-body mesh by inferring the abstract pose and shape parameters from visual content can obtain 3D bodies with realistic structures. However, the inferring process is highly non-linear and suffers from image-mesh misalignment, resulting in inaccurate reconstruction. In contrast, 3D keypoint estimation methods utilize the volumetric representation to achieve pixel-level accuracy but may predict unrealistic body structures. To address these issues, this paper presents a novel hybrid inverse kinematics solution, HybrIK, that integrates the merits of 3D keypoint estimation and body mesh recovery in a unified framework. HybrIK directly transforms accurate 3D joints to body-part rotations via twist-and-swing decomposition. The swing rotations are analytically solved with 3D joints, while the twist rotations are derived from visual cues through neural networks. To capture comprehensive whole-body details, we further develop a holistic framework, HybrIK-X, which enhances HybrIK with articulated hands and an expressive face. HybrIK-X is fast and accurate by solving the whole-body pose with a one-stage model. Experiments demonstrate that HybrIK and HybrIK-X preserve both the accuracy of 3D joints and the realistic structure of the parametric human model, leading to pixel-aligned whole-body mesh recovery. The proposed method significantly surpasses the state-of-the-art methods on various benchmarks for body-only, hand-only, and whole-body scenarios. Code and results can be found at \href{https://jeffli.site/HybrIK-X/}{https://jeffli.site/HybrIK-X/}.
\end{abstract}

}

\maketitle

\IEEEdisplaynontitleabstractindextext

\ifCLASSOPTIONpeerreview
\begin{center} \bfseries EDICS Category: 3-BBND \end{center}
\fi
%
\IEEEpeerreviewmaketitle


%
%
%

\IEEEraisesectionheading{\section{Introduction}}\label{sec:intro}

\IEEEPARstart{R}{ecovering} the whole-body 3D surface from visual content has a broad spectrum of applications. Advancements in parametric statistical human body shape models~\cite{anguelov2005scape,loper2015smpl,pavlakos2019expressive} have enabled the generation of realistic and animatable 3D meshes using only a small set of parameters. Despite the recent progress, recovering the 3D mesh by inferring the abstract pose and shape parameters is highly non-linear and still challenging.

Existing approaches can be divided into two categories: optimization-based and learning-based. Optimization-based approaches~\cite{guan2009estimating,bogo2016keep,pavlakos2019expressive} estimate the pose and shape of the human body through an iterative fitting process. The parameters of the statistical model are optimized to reduce the error between its 2D projection and 2D observations, e.g., 2D joint positions and silhouettes. However, this optimization problem is non-convex. Its solution can be time-consuming and its results are sensitive to the initialization. These challenges have shifted the research focus towards learning-based approaches. Learning-based approaches leverage parametric body models and employ neural networks to directly regress the pose parameters~\cite{hmr,spin,vibe,kocabas2021pare,pixie}. Nevertheless, the pose parameters represent relative rotations that lie in the $\mathcal{SO}(3)$ rotation group, which puts difficulties for neural networks to learn directly from RGB images. Consequently, the learned body mesh suffers from image-mesh misalignment.

\begin{figure}[t]
    \begin{center}
        \vspace{2mm}
        \includegraphics[width=\linewidth]{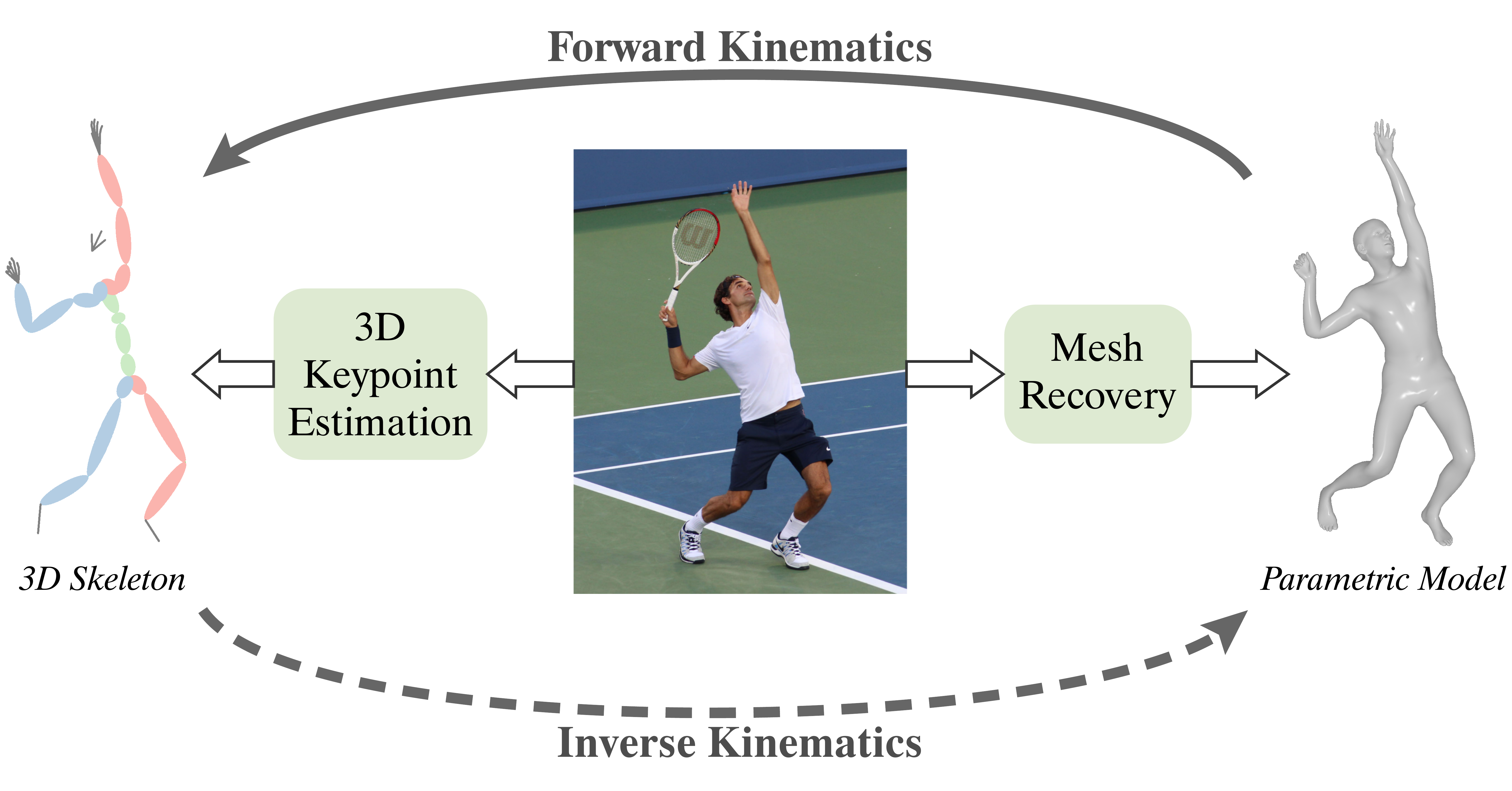}
    \end{center}
    \caption{\textbf{Closing the loop between the 3D skeleton and the parametric model via HybrIK/HybrIK-X.} 
    A 3D skeleton predicted by the neural network can be transformed into a parametric body mesh by inverse kinematics (IK) without loss of accuracy. The parametric body mesh can generate structural realistic 3D skeleton by forward kinematics (FK).}
    \label{fig:loop}
\end{figure}

Such a challenge prompts us to look into the field of 3D keypoint estimation. Previous 3D keypoint estimation approaches~\cite{coarse,integral,wang2020hmor} adopt volumetric heatmap as the target representation to learn 3D joint positions in the Cartesian coordinate system. The learned 3D joints can accurately align with the 2D RGB image. This inspires us to establish a collaboration between the 3D joints and the body mesh via forward kinematics (FK) and inverse kinematics (IK) (as illustrated in Fig.~\ref{fig:loop}). On the one hand, the accurate 3D joints can improve image-mesh alignment for mesh recovery. Since the body mesh is recovered from 3D joints, the recovered mesh can obtain pixel-aligned accuracy as long as the 3D joints are well-aligned with the image. On the other hand, the shape prior inherent in the parametric body model can be utilized to mitigate the issue of the unrealistic body structure in 3D keypoint estimation approaches. Since existing 3D keypoint estimation approaches lack explicit modeling of body bone length, they may predict unrealistic body structures like left-right asymmetry and abnormal proportions of limbs. If we can leverage the parametric body model, the presented human shape can better conform to the actual human body.

In this work, we present a hybrid analytical-neural inverse kinematics solution (HybrIK) to establish a collaboration between 3D keypoint estimation and whole-body mesh estimation. Inverse kinematics (IK) is used to find the corresponding body-part rotations from 3D body joints. This is an ill-posed problem due to the lack of a unique solution. The core of our approach is an innovative solution for this problem using twist-and-swing decomposition. Specifically, the rotation of a skeleton part is decomposed into \textit{twist} and \textit{swing}, i.e., a longitudinal rotation and an in-plane rotation. The unique solution of the body-part rotations is composited iteratively along the kinematic tree by analytically calculating \textit{swing} rotations from 3D joints and using a neural network to predict \textit{twist} rotations from visual cues.

This IK framework is extended as HybrIK-X for articulated hand and expressive face reconstruction, improving the fine-grained image-mesh alignment with a new backward-updated solution. Unlike previous approaches~\cite{rong2021frankmocap,pixie,zhang2022pymaf}, HybrIK-X gets rid of separated expert models and recovers the whole-body mesh with a one-stage network, resulting in improved efficiency and reduced computational resources. The robustness of 3D keypoints estimation against occlusions and truncations is enhanced by using a new regression approach to infer the truncated body parts. Furthermore, we exploit the body structure from the parametric human body model to alleviate the depth ambiguity in predicting camera parameters and estimate a more stable human motion.

A critical characteristic of our approach is that the estimated mesh is inherently aligned with the 3D skeleton, without the need for additional optimization procedures in the previous approaches~\cite{bogo2016keep,pavlakos2019expressive,spin}. We conduct comprehensive experiments on various benchmarks for body-only, hand-only, and whole-body scenarios, including 3DPW~\cite{3dpw}, Human3.6M~\cite{h36m}, and MPI-INF-3DHP~\cite{3dhp}, FreiHAND~\cite{zimmermann2019freihand}, HO3D~\cite{hampali2020honnotate}, and AGORA~\cite{patel2021agora}. HybrIK and HybrIK-X show pixel-aligned accuracy and significantly outperform state-of-the-art approaches.

The contributions of our approach can be summarized as follows:
\begin{itemize}
    \item We propose a novel whole-body mesh recovery framework that uses a hybrid analytical-neural IK algorithm to convert accurate 3D joints to pixel-aligned body meshes.
    \item Our approach closes the loop between the 3D skeleton and the parametric model. It improves the image-mesh alignment for body mesh recovery and addresses the unrealistic body structure problem of 3D keypoint estimation approaches at the same time.
    \item Our approach achieves state-of-the-art performance across various body-only, hand-only, and whole-body benchmarks.
\end{itemize}

A preliminary version of this work was accepted in CVPR 2021~\cite{li2021hybrik}. This paper extends the previous work in the following ways. First, we extend the IK framework to whole-body mesh recovery with a backward-updated IK solution and an efficient one-stage model. Second, we enhance the robustness of our framework to occlusions and truncations by a new regression paradigm. Third, we propose a structure-aware cycle for the mitigation of depth ambiguity, thereby providing a more stable camera parameters estimation. Fourth, additional quantitative comparisons on various benchmarks for body-only, hand-only, and whole-body scenarios demonstrate the effectiveness and generalization of the proposed IK framework. Finally, we conduct further ablation studies to investigate and analyze our framework.

\section{Related Work}\label{sec:related}

\subsection{3D Keypoint Estimation}
Many studies formulates 3D human pose estimation as the problem of locating the 3D joints of the human body. Previous work can be divided into two categories: single-stage and two-stage approaches.
Single-stage approaches~\cite{pavlakos2017coarse,rogez2017lcr,mehta2017monocular,zhou2017towards,mehta2018single,sun2018integral,moon2019camera,wang2020hmor} directly estimate the 3D joint locations from the input image. Various representations are developed, including 3D heatmap~\cite{pavlakos2017coarse}, location-map~\cite{mehta2017monocular}, and 2D heatmap + $z$ regression~\cite{zhou2017towards}. Two-stage approaches first estimate 2D pose and then lift them to 3D joint locations by a learned dictionary of 3D skeleton~\cite{akhter2015pose,ramakrishna2012reconstructing,tung2017adversarial,sanzari2016bayesian,zhou2016sparse,zhou2016sparseness} or regression~\cite{park20163d,yasin2016dual,moreno20173d,fang2017learning,martinez2017simple,sun2017compositional}. Two-stage approaches highly rely on accurate 2D pose estimators, which have achieved impressive performance through the combination of a powerful backbone network~\cite{vgg,resnet,hourglass,pang2019deep,pang2020complex} and the 2D heatmap.

These privileged forms of supervision contribute to the recent performance leaps of 3D keypoint estimation. However, the human structural information is modeled implicitly by the neural network, which can not ensure the output 3D skeletons are realistic. Our approach combines the advantages of both the 3D skeleton and parametric model to predict accurate and realistic human pose and shape.

\subsection{Model-based 3D Body Pose and Shape Estimation} Prior work on the model-based 3D pose and shape estimation uses parameters of the statistical body model~\cite{anguelov2005scape,loper2015smpl,pavlakos2019expressive} as the output target because they capture the statistics prior of body shape. Compared with the model-free methods~\cite{varol2018bodynet,kolotouros2019convolutional,moon2020i2l}, the model-based methods directly predict controllable body mesh, which can facilitate many downstream tasks for both computer graphics and computer vision. Bogo et al.~\cite{bogo2016keep} propose SMPLify, a fully automatic approach, without manual user intervention~\cite{sigal2008combined,guan2009estimating}. This optimization paradigm was further extended with silhouette cues~\cite{lassner2017unite}, volumetric grids~\cite{varol2018bodynet}, multiple people~\cite{zanfir2018monocular}, and whole-body parametric models~\cite{pavlakos2019expressive}.

With the advances in deep learning networks, there are increasing studies that focus on learning-based methods~\cite{pavlakos2018learning,omran2018neural,hmr,spin,vibe,kocabas2021pare,zhang2021pymaf,sun2021monocular,iqbal2021kama,lin2021end}, using a deep network to estimate the pose and shape parameters. Since the mapping from RGB image to body shape and body-part rotation is hard to learn, many studies use intermediate representations to alleviate this problem, such as keypoints and silhouettes~\cite{pavlakos2018learning}, semantic part segmentation~\cite{omran2018neural}, and 2D heatmap input~\cite{tung2017self}. Kanazawa et al.~\cite{hmr} use an adversarial prior and an iterative error feedback (IEF) loop to reduce the difficulty of regression. Arnab et al.~\cite{arnab2019exploiting} and Kocabas et al.~\cite{vibe} exploit temporal context, while Guler et al.~\cite{guler2019holopose} use a part-voting expression and test-time post-processing to improve the regression network. Kolotouros et al.~\cite{spin} leverage the optimization paradigm to provide extra 3D supervision from unlabeled images. Zhang et al.~\cite{zhang2020object} propose to use saliency maps to infer occluded bodies. PARE~\cite{kocabas2021pare} uses part-based attention to improve body-part regression. PyMAF~\cite{zhang2021pymaf} uses a mesh-aligned feedback loop to exploit locally aligned features.

In this work, we address this challenging learning problem by a transformation from the pixel-aligned 3D joints to the body-part rotations.

\subsection{Whole-body Mesh Recovery} Numerous prior studies solve body, face~\cite{jackson2017large,feng2018joint,tewari2018self,sanyal2019learning,feng2021learning}, and hand meshes~\cite{zhang2019end,baek2019pushing,boukhayma20193d,hasson2019learning,kulon2019single,ge20193d,moon2020interhand2,kulon2020weakly,zhang2021interacting,rong2021monocular,li2022interacting} separately. Recent studies start using whole-body statistical models~\cite{joo2018total,pavlakos2019expressive,xu2020ghum} to jointly recover the 3D surface of the body, face, and hands.

Pavlakos et al.~\cite{pavlakos2019expressive} propose to estimate whole-body mesh by automatically fitting the SMPL-X model~\cite{pavlakos2019expressive} to the 2D body, face, and hand keypoints estimated by off-the-shelf whole-body keypoint estimators~\cite{openpose,alphapose}. Xiang et al.~\cite{xiang2019monocular} propose to learn 2D keypoints and part orientation fields for model fitting. Xu et al.~\cite{xu2020ghum} fit GHUM with a reprojection error and a semantic body-part alignment error under anatomical joint angle limits. Similar to body-only pose estimation, optimization-based methods are slow and sensitive to initialization.

Learning-based methods tackle the above limitations by directly regressing the pose and shape parameters from the input image. ExPose~\cite{choutas2020monocular} uses three expert sub-networks to estimate body, face, and hands parameters separately. The body expert first estimates the body pose and the rough poses for the face and hands. Then the face and hand experts estimate the refined part poses from the cropped local image. Finally, the whole-body mesh is reconstructed by merging the results from three experts. Follow-up studies follow the same paradigm of ExPose~\cite{choutas2020monocular}. They use separate experts to handle body, face, and hand poses and merge them into a holistic whole-body pose. FrankMocap~\cite{rong2021frankmocap} integrates the results from three experts through an integration network to approximate the optimization to better align the wrist and hand poses. Zhou et al.~\cite{zhou2021monocular} propose to regress the body and hand poses from the detected keypoints. PIXIE~\cite{feng2021collaborative} uses an attention-based moderator network to integrate body, face, and hand results from experts. Hand4Whole~\cite{moon2022accurate} introduces to learn the wrist rotations from the hand keypoints. PyMAF-X~\cite{zhang2022pymaf} proposes an adaptive integration module to better integrate the elbow and wrist poses.

Although using separate experts can benefit network training by using high-resolution input and multiple data sources, it costs higher computational complexity and lacks holistic information to integrate results from different experts. Unlike previous methods, we propose a one-stage paradigm that directly estimates the whole-body mesh. Our one-stage method gets rid of the time-consuming separated experts and the entire model is trained in an end-to-end manner.

\subsection{Body-part Rotation in Pose Estimation} The core of our approach is to calculate the relative rotation of human body parts through a hybrid IK process. There are several studies that estimate the rotations in the 3D pose estimation literature. Zhou et al.~\cite{zhou2016deep} use the network to predict the rotation angle of each body joint, followed by an FK layer to generate the 3D joint coordinates. Pavllo et al.~\cite{pavllo2018quaternet} switch to quaternions, while Yoshiyasu et al.~\cite{yoshiyasu2018skeleton} directly predict the $3\times3$ rotation matrices. Mehta et al.~\cite{mehta2019xnect} first estimate the 3D joints and then use a fitting procedure to find the rotation Euler angles. Previous approaches are either limited to a hard-to-learn problem or require an additional fitting procedure. Our approach recovers the body-part rotation from 3D joint locations in a direct, accurate and feed-forward manner.

\begin{figure*}[t]
    \begin{center}
        \includegraphics[width=\linewidth]{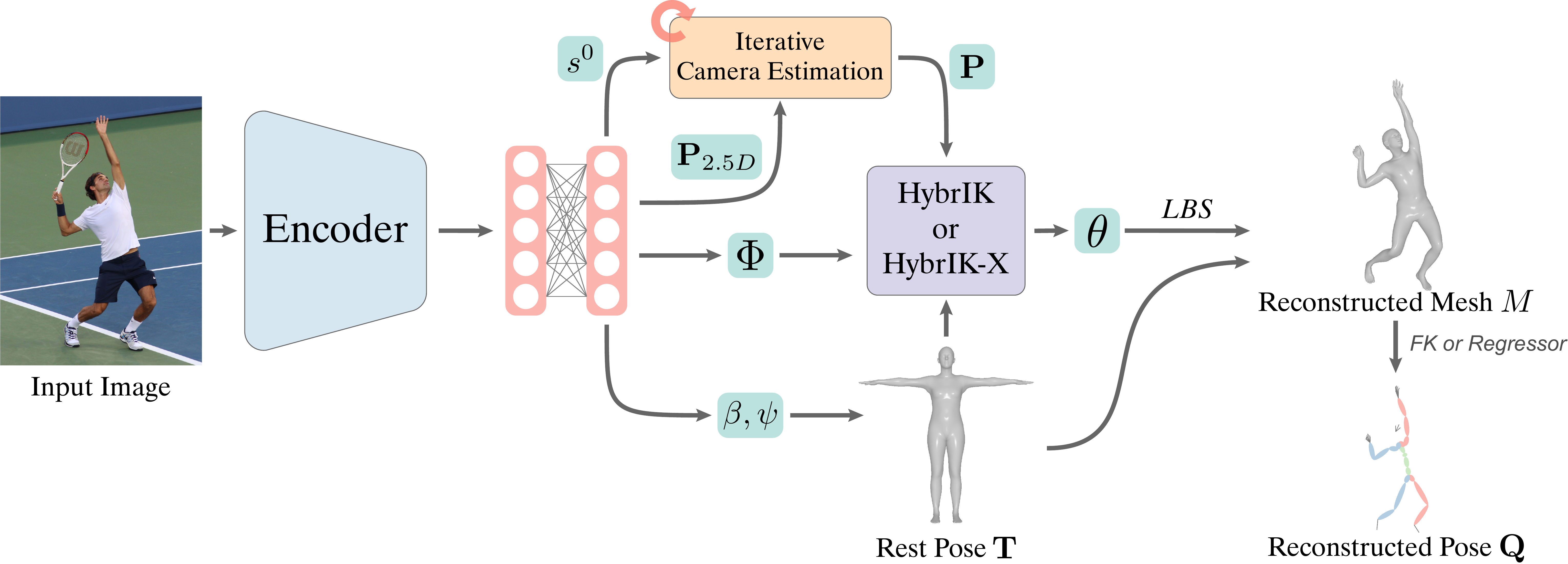}
    \end{center}
    \caption{\textbf{Overview of the proposed inverse kinematics framework.} 2.5D joints $\mathbf{P}_{2.5D}$, shape parameters $\beta$, expression parameters $\psi$, \textit{twist} angles $\Phi$, and the initial camera parameter $s^0$ are learned from the visual cues through neural networks. The 2.5D joints $\mathbf{P}_{2.5D}$ and the initial camera parameter $s^0$ are fed into our iterative camera estimation algorithm to obtain the estimated camera and the back-projected 3D joints $\mathbf{P}$. These results are then sent to the HybrIK process to solve the body-part rotations, i.e., the pose parameters $\theta$. Finally, with the pose and shape parameters, we can obtain the reconstructed body mesh $M$ via linear blend skinning (LBS), and the reconstructed pose ${\mathbf{Q}}$ via a further FK process or linear regression.
    }
    \label{fig:pipeline}
\end{figure*}

\subsection{Inverse Kinematics Process} The inverse kinematics (IK) problem has been extensively studied during recent decades. Numerical solutions~\cite{balestrino1984robust,wolovich1984computational,girard1985computational,klein1983review,wampler1986manipulator,buss2005selectively} are simple ways to implement the IK process, but they suffer from time-consuming iterative optimization.
Heuristic methods are efficient solutions to the IK problem. For example, CDC~\cite{luenberger1984linear}, FABRIK~\cite{aristidou2011fabrik}, and IK-FA~\cite{rokbani2015ik} have a low computational cost for each heuristic iteration. In some special cases, there exist analytical solutions to the IK problem. Tolani et al.~\cite{tolani2000real} propose a reliable algorithm by the combination of analytical and numerical methods. Kallmann et al.~\cite{kallmann2008analytical} solve the IK for arm linkage, i.e., a three-joint system. Recently, researchers have been interested in using neural networks to solve the IK problem in robotic control~\cite{csiszar2017solving}, motion retargeting~\cite{villegas2018neural}, and hand pose estimation~\cite{mueller2017real,kokic2019learning}.

In this work, we combine the interpretable characteristic of the analytical solution and the flexibility of the neural network, introducing a feed-forward hybrid IK algorithm with twist-and-swing decomposition. Twist-and-swing decomposition is first introduced by Baerlocher et al.~\cite{baerlocher2001parametrization}. The twist angles are limited based on the particular body joint. In our work, the twist angles are estimated by a neural network, which is more flexible and can be generalized to all body joints. Compared with previous analytical solutions~\cite{kallmann2008analytical} designed for specific joint linkage, our algorithm can be applied to the entire body skeleton in a direct and differentiable manner.

\section{Method}

In this section, we present our hybrid analytical-neural inverse kinematics solution for whole-body mesh recovery (Fig.~\ref{fig:pipeline}). First, in \S\ref{sec:bg}, we briefly describe the forward kinematics process, the inverse kinematics process, and the SMPL/SMPL-X model. In \S\ref{sec:tas-ik}, we introduce the proposed inverse kinematics solution, HybrIK, for body-only mesh recovery, and HybrIK-X, for whole-body mesh recovery. Then, in \S\ref{sec:learning}, we present the overall learning framework to estimate the pixel-aligned whole-body mesh and realistic 3D skeleton. Finally, we provide the necessary implementation details in \S\ref{sec:implement}.

\subsection{Preliminary}\label{sec:bg}

\textbf{Forward Kinematics.} Forward kinematics (FK) for human pose usually refers to the process of computing the reconstructed pose $\mathbf{Q} = \{{q}_k\}_{k=1}^{K}$, with the rest pose template $\mathbf{T} = \{{t}_k\}_{k=1}^{K}$ and the relative rotations $\mathbf{R} = \{R_{{\mathtt{pa}(k)},{k}}\}_{k=1}^{K}$ as input:
\begin{equation}
    \mathbf{Q} = \textit{FK}(\mathbf{R}, \mathbf{T}),
    \label{eq:fk}
\end{equation}
where $K$ is the number the body joints, $q_k \in \mathbb{R}^3$ denotes the reconstructed 3D location of the $k$-th joint, $t_k \in \mathbb{R}^3$ denotes the $k$-th joint location of the rest pose template, $\mathtt{pa}(k)$ returns the parent's index of the $k$-th joint, and $R_{{\mathtt{pa}(k)},{k}}$ is the relative rotation of $k$-th joint with respect to its parent joint. FK can be performed by recursively rotating the template body part from the root joint to the leaf joints:
\begin{equation}
    q_k = R_k(t_{k} - t_{\mathtt{pa}(k)}) + q_{\mathtt{pa}(k)},
    \label{eq:fk-step}
\end{equation}
where $R_{k} \in \mathbb{SO}(3)$ is the global rotation of the $k$-th joint with respect to the canonical rest pose space. The global rotation can be calculated recursively:
\begin{equation}
    R_k = R_{\mathtt{pa}(k)} R_{\mathtt{pa}(k), k}.
    \label{eq:r-step}
\end{equation}
For the root joint that has no parent, we have $q_0 = t_0$.

~

\noindent
\textbf{Inverse Kinematics.} Inverse kinematics (IK) is the reverse process of FK, computing relative rotations $\mathbf{R}$ that can generate the desired locations of input body joints $\mathbf{P} = \{p_k\}_{k=1}^{K}$. This process can be formulated as:
\begin{equation}
    \mathbf{R} = \textit{IK}(\mathbf{P}, \mathbf{T}),
\end{equation}
where $p_k$ denotes the $k$-th joint of the input pose. Ideally, the resulting rotations should satisfy the following condition:
\begin{equation}
    p_k - p_{\mathtt{pa}(k)} = R_k(t_k - t_{\mathtt{pa}(k)}) \quad \forall 1 \leq k \leq K.
    \label{eq:ik-cond}
\end{equation}
Similar to the FK process, we have $p_0 = t_0$ for the root joint that has no parent. While the FK problem is well-posed, the IK problem is ill-posed because there are either no solutions or too many solutions to fulfill the target joint locations.

~

\noindent
\textbf{SMPL and SMPL-X Models.} In this work, we employ the SMPL~\cite{loper2015smpl} parametric model for body-only mesh recovery and the expressive SMPL-X~\cite{pavlakos2019expressive} model for whole-body mesh recovery. SMPL-X is the whole-body extension of SMPL. It allows us to use shape parameters, expression parameters and pose parameters to control the whole-body mesh. The shape parameters $\beta \in \mathbb{R}^{200}$ are parameterized by the first $200$ principal components of a linear shape space, learned from registered CAESAR~\cite{robinette2002civilian} scans. The expression parameters $\psi \in \mathbb{R}^{50}$ are coefficients of a low-dimensional linear space. The pose parameters $\theta$ are modelled by relative 3D rotations of $K = 55$ joints, $\theta = (\theta_1,\theta_2,\cdots,\theta_K)$, consisting of body, jaw and hand poses. SMPL-X provides a differentiable skinning function $\mathcal{M}(\theta, \beta, \psi)$ that takes pose parameters $\theta$, shape parameters $\beta$, and expression parameters $\psi$ as input and outputs a triangulated mesh $M \in \mathbb{R}^{N \times 3}$ with $N=10475$ vertices. The reconstructed 3D joints ${\mathbf{Q}}_{\textit{smpl-x}}$ can be conveniently obtained by the FK process, i.e., ${\mathbf{Q}}_{\textit{smpl-x}} = \textit{FK}(\mathbf{R}, \mathbf{T})$. Also, other format of joints can be obtained by a linear combination of the mesh vertices through a linear regressor $W$, i.e., ${\mathbf{Q}} = WM$.

\subsection{Hybrid Analytical-Neural Inverse Kinematics}\label{sec:tas-ik}
Estimating the human body mesh by directly regressing the body part rotations is too difficult~\cite{hmr,spin,vibe}. In this paper, we propose a hybrid analytical-neural inverse kinematics solution that use 3D keypoints to recover 3D body mesh. Since the IK problem is ill-posed, we cannot uniquely determine the relative rotation just by the 3D joints. Here, we first decompose the original rotation into \textit{twist} and \textit{swing}. The 3D joints are utilized to calculate the \textit{swing} rotation analytically, and we exploit the visual cues by a neural network to estimate the 1-DoF \textit{twist} rotation. In our IK algorithm, the body part rotations are solved recursively along the kinematic tree.

\begin{figure}[tb]
    \begin{center}
        \includegraphics[width=0.95\linewidth]{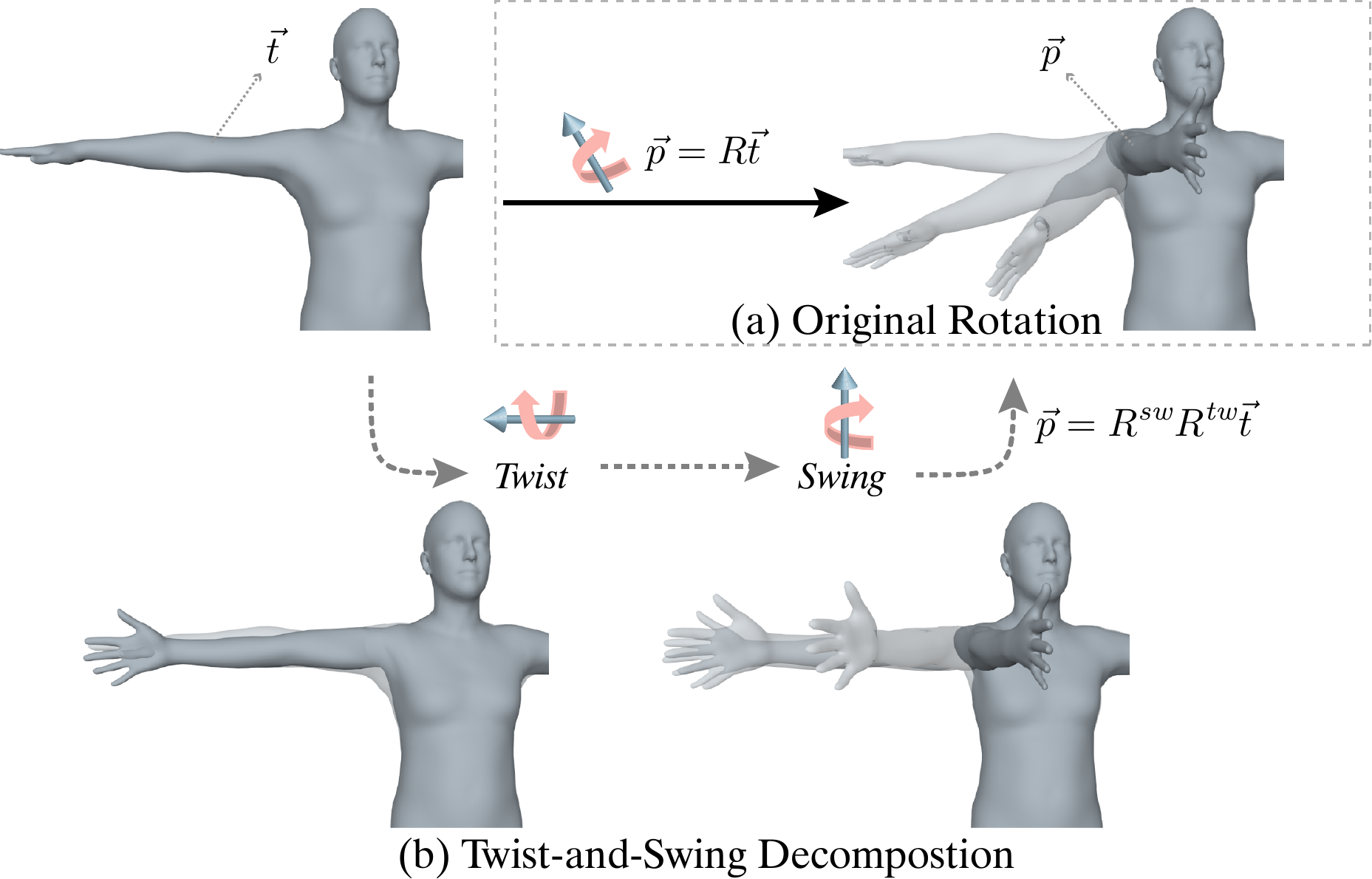}
    \end{center}
    \caption{\textbf{Illustration of the twist-and-swing decomposition.} (a) The original rotation moves the right hand to the front and turns the palm to the left in one step. (b) With twist-and-swing decomposition, the rotation can be divided into two steps: First, turn the palm $90^\circ$ by the \textit{twist} rotation, and then move the entire hand to the front by the \textit{swing} rotation.}
    \label{fig:pat}
\end{figure}

\subsubsection{Twist-and-Swing Decomposition.}
In the conventional analytical IK formulation, some body joints are usually assigned lower degree-of-freedom (DoFs) to simplify the problem, e.g., $1$ or $2$ DoFs~\cite{kofinas2013complete,tolani2000real,kallmann2008analytical}. In this work, we consider a general case where each body joint is assumed to have full $3$ DoFs. As illustrated in Fig.~\ref{fig:pat}, a rotation $R \in \mathbb{SO}(3)$ can be decomposed into a \textit{twist} rotation $R^{\textit{tw}}$ and a \textit{swing} rotation $R^{\textit{sw}}$. Given the start template body-part vector $\vec{t}$ and the target vector $\vec{p}$, the solution process of $R$ can be formulated as:
\begin{equation}
    R = \mathcal{D}(\vec{p}, \vec{t}, \phi) = \mathcal{D}^{\textit{sw}}(\vec{p}, \vec{t})\mathcal{D}^{\textit{tw}}(\vec{t}, \phi) = R^{\textit{sw}}R^{\textit{tw}},
    \label{eq:dec}
\end{equation}
where $\phi$ is the \textit{twist} angle that estimated by a neural network, $\mathcal{D}^{\textit{sw}}(\cdot)$ is a closed-form solution of the \textit{swing} rotation, and $\mathcal{D}^{\textit{tw}}(\cdot)$ transforms $\phi$ to the \textit{twist} rotation. Here, $R$ should satisfy the condition in Eq.~\ref{eq:ik-cond}, i.e., $\vec{p} = R \vec{t}$.

~

\noindent
\textbf{Swing Rotation.}
The \textit{swing} rotation has the axis $\vec{n}$ that is perpendicular to $\vec{t}$ and $\vec{p}$. Therefore, it can be formulated as:
\begin{equation}
    \vec{n} = \frac{\vec{t}\times\vec{p}}{\|~ \vec{t}\times\vec{p} ~\|},
    \label{eq:swing-axis}
\end{equation}
and the \textit{swing} angle $\alpha$ satisfies:
\begin{equation}
    \cos{\alpha} = \frac{\vec{t}\cdot\vec{p}}{\|\vec{t}\| \|\vec{p}\|}, \quad
    \sin{\alpha} = \frac{\|~ \vec{t}\times\vec{p} ~\|}{\|\vec{t}\| \|\vec{p}\|}.
    \label{eq:swing-angle}
\end{equation}
Hence, the closed-form solution of the \textit{swing} rotation $R^{\textit{sw}}$ can be derived by the \textit{Rodrigues formula}:
\begin{equation}
    R^{\textit{sw}} = \mathcal{D}^{\textit{sw}}(\vec{p}, \vec{t}) = \mathcal{I} + \sin{\alpha}[\vec{n}]_{\times} + (1 - \cos{\alpha})[\vec{n}]_{\times}^2,
    \label{eq:swing-rod}
\end{equation}
where $[\vec{n}]_{\times}$ is the skew symmetric matrix of $\vec{n}$ and $\mathcal{I}$ is the $3\times3$ identity matrix.

\noindent
\textbf{Twist Rotation.}
The \textit{twist} rotation is rotating around $\vec{t}$ itself. Thus, with $\vec{t}$ itself the axis and $\phi$ the angle, we can determine \textit{twist} rotation $R^{tw}$:
\begin{equation}
    R^{\textit{tw}} = \mathcal{D}^{\textit{tw}}(\vec{t}, \phi) = \mathcal{I} + \frac{\sin{\phi}}{\| \vec{t} \|}[\vec{t}]_{\times} + \frac{(1 - \cos{\phi})}{\| \vec{t} \|^2}[\vec{t}]_{\times}^2,
    \label{eq:twist-rod}
\end{equation}
where $[\vec{t}]_{\times}$ is the skew symmetric matrix of $\vec{t}$.

Note that function $\mathcal{D}^{\textit{sw}}$ and $\mathcal{D}^{\textit{tw}}$ are fully differentiable, which allows us to integrate the twist-and-swing decomposition into the training process.
Although we need a neural network to learn the \textit{twist} angle, the difficulty of learning is significantly reduced. Compared with previous work~\cite{hmr,spin,vibe} that directly regresses the 3-DoF rotation, we regress the \textit{twist} angle that is only a 1-DoF variable.
Moreover, due to the physical limitation of the human body, the \textit{twist} angle has a small range of variation. Therefore, it is much easier for the networks to learn the mapping function. We further analyze its variation in \S\ref{sec:ablation}. 

\subsubsection{Body-only Inverse Kinematics}

{
\begin{algorithm}[t]
    \label{alg:naive-hybrik}
    \caption{Naive HybrIK}
    \KwIn{$\mathbf{P}$, $\mathbf{T}$, $\Phi$}
    \KwOut{$\mathbf{R}$}
    Determine $R_{0}$\;
    \For{$k$ along the kinematic tree}
    {
        $\vec{p}_k \leftarrow R_{\mathtt{pa}(k)}^{-1}(p_k - p_{\mathtt{pa}(k)})$\;
        $\vec{t}_k \leftarrow (t_k - t_{\mathtt{pa}(k)})$\;
        $R^{\textit{sw}}_{\mathtt{pa}(k),k} \leftarrow \mathcal{D}^{\textit{sw}}(\vec{p}_k, \vec{t}_k)$\;
        $R^{\textit{tw}}_{\mathtt{pa}(k),k} \leftarrow \mathcal{D}^{\textit{tw}}(\vec{t}_k, \phi_k)$\;
        $R_{\mathtt{pa}(k),k} \leftarrow R^{\textit{sw}}_{\mathtt{pa}(k),k} R^{\textit{tw}}_{\mathtt{pa}(k),k}$\;
    }
\end{algorithm}
}

\noindent
\textbf{Naive HybrIK.} Using the twist-and-swing decomposition, the IK process can be performed recursively along the kinematic tree like the FK process. First of all, we need to determine the global root rotation $R_{0}$, which has a closed-form solution using the locations of $\mathtt{spine}$, $\mathtt{left~hip}$, $\mathtt{right~hip}$ and Singular Value Decomposition (SVD). Detailed mathematical proof is provided in appendix \S\ref{sec:svd}. Then, in each step, e.g., the $k$-th step, we assume the rotation of the parent joint $R_{\mathtt{pa}(k)}$ is known. Hence, we can reformulate Eq.~\ref{eq:ik-cond} with Eq.~\ref{eq:r-step} as:
\begin{equation}
    R_{\mathtt{pa}(k)}^{-1}(p_k - p_{\mathtt{pa}(k)}) = R_{\mathtt{pa}(k), k}(t_k - t_{\mathtt{pa}(k)}).
\end{equation}
Let $\vec{p}_k = R_{\mathtt{pa}(k)}^{-1}(p_k - p_{\mathtt{pa}(k)})$ and $\vec{t}_k = (t_k - t_{\mathtt{pa}(k)})$, we can solve the relative rotation via Eq.~\ref{eq:dec}:
\begin{equation}
    R_{\mathtt{pa}(k), k} = \mathcal{D}(\vec{p}_k, \vec{t}_k, \phi_k),
\end{equation}
where $\phi_k$ is the network-predicted $\textit{twist}$ angle for the $k$-th joint. The set of \textit{twist} angle is denoted as $\Phi = \{\phi_k\}_{k=1}^K$. Since the rotation matrices are orthogonal, their inverse equals to their transpose, i.e., $R_{\mathtt{pa}(k)}^{-1} =  R_{\mathtt{pa}(k)}^\mathrm{T}$, which keeps the solving process differentiable.

The whole process is named Naive HybrIK and summarized in Alg.~\ref{alg:naive-hybrik}.
Note that we solve the relative rotation $R_{\mathtt{pa}(k), k}$ instead of the global rotation $R_k$.
The reason is that if we directly decompose the global rotation, the resulting \textit{twist} angle will depend on all ancestors' rotations along the kinematic tree, which increases the variation of the distal limb joints and makes it difficult for the network to learn.

~

\noindent
\textbf{Adaptive HybrIK.} Although the Naive HybrIK process seems effective, it follows an unstated hypothesis: $\| p_k - p_{\mathtt{pa}(k)} \| = \| t_k - t_{\mathtt{pa}(k)} \|$. Otherwise, there is no solution for Eq.~\ref{eq:ik-cond}. Unfortunately, in our case, the body-parts predicted by the 3D keypoint estimation method are not always consistent with the rest pose template. In Naive HybrIK, Eq.~\ref{eq:dec} can still be solved because the condition is turned into:
\begin{equation}
    p_k - p_{\mathtt{pa}(k)} = R_k(t_k - t_{\mathtt{pa}(k)}) + \vec{\epsilon}_k,
    \label{eq:ik-cond-2}
\end{equation}
where $\vec{\epsilon}_k$ denotes the error in the $k$-th step, which has the same direction of $p_k - p_{\mathtt{pa}(k)}$ and $\| \vec{\epsilon}_k \| = \lvert \| p_k - p_{\mathtt{pa}(k)} \| - \| t_k - t_{\mathtt{pa}(k)} \| \rvert$. 
To analyze the reconstruction error, we compare the difference between the input pose $\mathbf{P}$ and the reconstructed pose $\mathbf{Q}$:
\begin{equation}
    \| \mathbf{P} - \mathbf{Q} \| \Leftrightarrow \sum_{k=1}^K \| p_k - q_k \|,
\end{equation}
where $\mathbf{Q} = \textit{FK}(\mathbf{R}, \mathbf{T}) = \textit{FK}(\textit{IK}(\mathbf{P}, \mathbf{T}), \mathbf{T})$. Combining Eq.~\ref{eq:fk-step} and Eq.~\ref{eq:ik-cond-2}, we have:
\begin{equation}
    \begin{aligned}
    p_k - q_k &= p_{\mathtt{pa}(k)} - q_{\mathtt{pa}(k)} + \vec{\epsilon}_k \\
    &= p_{\mathtt{pa}^2(k)} - q_{\mathtt{pa}^2(k)} + \vec{\epsilon}_{\mathtt{pa}(k)} + \vec{\epsilon}_k \\
    &= \ldots =  \sum_{i \in \mathcal{A}(k)} \vec{\epsilon}_i,
    \end{aligned}
    \label{eq:naive-error}
\end{equation}
where $\mathtt{pa}^2(k)$ denotes the parent index of the $\mathtt{pa}(k)$-th joint, and $\mathcal{A}(k)$ denotes the set of ancestors of the $k$-th joint. That means the difference between the input joint $p_k$ and the reconstructed joint $q_k$ will accumulate along the kinematic tree, which brings more uncertainty to the distal joint.

\begin{figure}[tb]
    \begin{center}
        \includegraphics[width=0.95\linewidth]{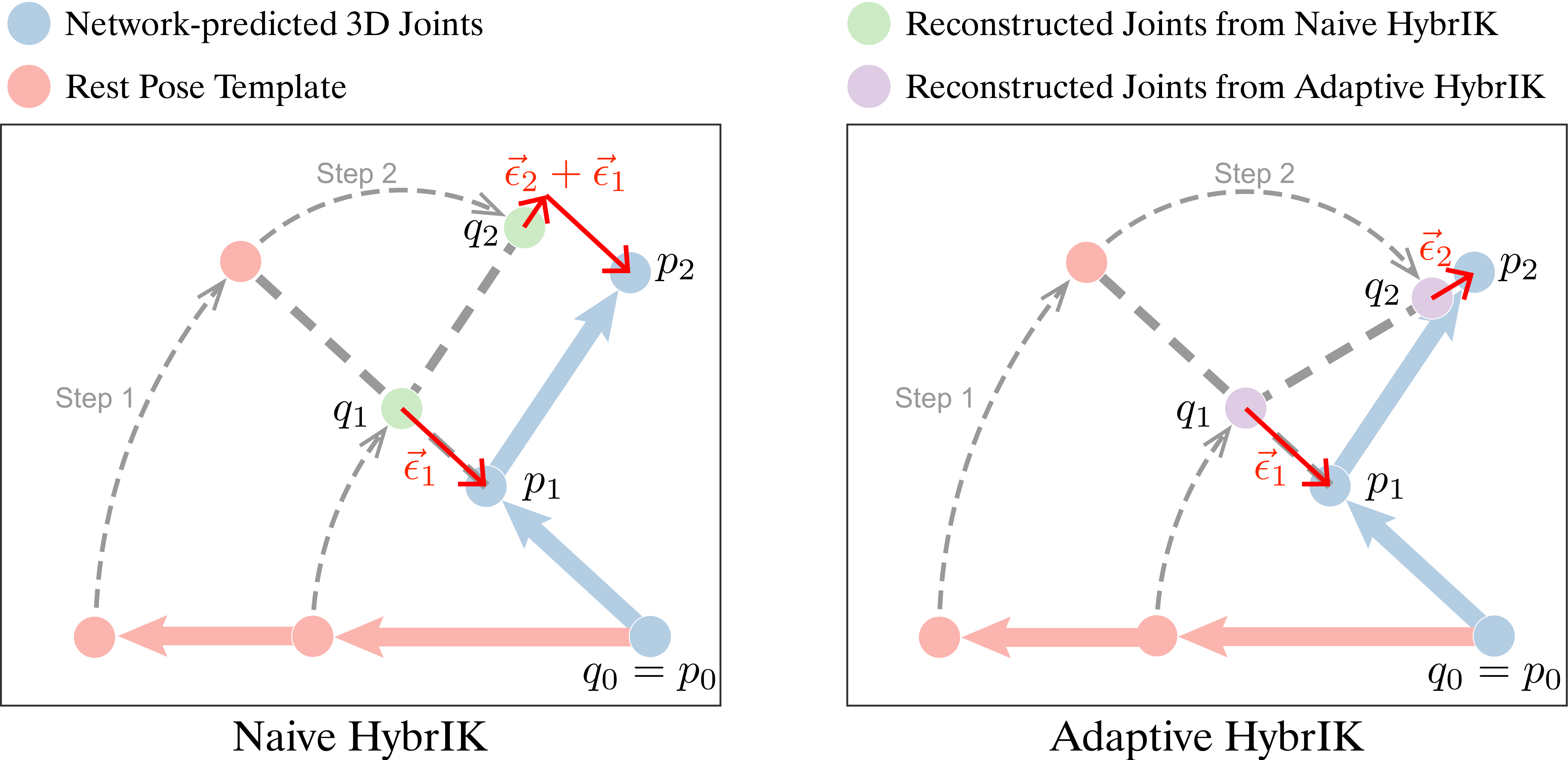}
    \end{center}
    \caption{\textbf{Example of the reconstruction error.} The joints in the rest pose are rotated to $q_1$ and $q_2$ by two steps. In the first step, due to the bone-length inconsistency, the reconstruction error is $\vec{\epsilon}_1$. In the second step, Naive HybrIK takes $p_2 - p_1$ as the target direction, resulting in the accumulation of error $\vec{\epsilon}_1 + \vec{\epsilon}_2$. Instead, Adaptive HybrIK selects the reconstructed joint $q_1$ to form the target direction $p_2 - q_1$, which reduces the error to only $\vec{\epsilon}_2$.}
    \label{fig:error}
\end{figure}

To address this error accumulation problem, we further propose Adaptive HybrIK. In Adaptive HybrIK, the target vector is adaptively updated with the newly reconstructed parent joints.
Let $\vec{p}_k = R_{\mathtt{pa}(k)}^{-1}(p_k - q_{\mathtt{pa}(k)})$ and $\vec{t}_k$ the same as the one in the naive solution. In this way, the condition in Adaptive HybrIK can be formulated as:
\begin{equation}
    p_k - q_{\mathtt{pa}(k)} = R_k(t_k - t_{\mathtt{pa}(k)}) + \vec{\epsilon}_k.
    \label{eq:ik-cond-3}
\end{equation}
Therefore, we have:
\begin{equation}
    \begin{aligned}
    &{p_k} - q_{\mathtt{pa}(k)} = q_k - q_{\mathtt{pa}(k)} + \vec{\epsilon}_k \\
    \Rightarrow &\quad p_k - q_k = \vec{\epsilon}_k.
    \end{aligned}
\end{equation}
Compared to the naive solution (Eq.~\ref{eq:naive-error}), the reconstructed error of the adaptive solution only depends on the current joint position. As illustrated in Fig.~\ref{fig:error}, in Naive HybrIK, once the parent joint is out of position, its children will continue this mistake. Instead, in Adaptive HybrIK, the solved relative rotation is always pointing towards the target joint and tries to reduce the error. We conduct empirical experiments in \S\ref{sec:ablation} to validate its robustness. The whole process of Adaptive HybrIK is summarized in Alg.~\ref{alg:adaptive-hybrik}.

\begin{algorithm}[t]
    \label{alg:adaptive-hybrik}
    \caption{Adaptive HybrIK}
    \KwIn{$\mathbf{P}$, $\mathbf{T}$, $\Phi$}
    \KwOut{$\mathbf{R}$}
    Determine $R_{0}$\;
    \For{$k$ along the kinematic tree}
    {
        $q_{\mathtt{pa}(k)} \leftarrow R_{\mathtt{pa}(k)}(t_{\mathtt{pa}(k)} - t_{\mathtt{pa}^2(k)}) + q_{\mathtt{pa}^2(k)}$ \;
        $\vec{p}_k \leftarrow R_{\mathtt{pa}(k)}^{-1}(p_k - q_{\mathtt{pa}(k)})$\;
        $\vec{t}_k \leftarrow (t_k - t_{\mathtt{pa}(k)})$\;
        $R^{\textit{sw}}_{\mathtt{pa}(k),k} \leftarrow \mathcal{D}^{\textit{sw}}(\vec{p}_k, \vec{t}_k)$\;
        $R^{\textit{tw}}_{\mathtt{pa}(k),k} \leftarrow \mathcal{D}^{\textit{tw}}(\vec{t}_k, \phi_k)$\;
        $R_{\mathtt{pa}(k),k} \leftarrow R^{\textit{sw}}_{\mathtt{pa}(k),k} R^{\textit{tw}}_{\mathtt{pa}(k),k}$\;
    }
\end{algorithm}

\subsubsection{Whole-body Inverse Kinematics}
Adaptive HybrIK is accurate for body-only mesh recovery. It reduces the accumulated errors by adaptively updating the parent joints in a feedforward process. Nevertheless, extending it to whole-body mesh recovery is nontrivial. Although adaptive HybrIK tries to minimize the error in each step, the error won't be totally eliminated since we cannot fix the erroneous positions of the ancestor joints. As long as the bone length calculated from keypoints is different from the SMPL-X model, misalignment is inevitable. This misalignment is particularly pronounced when considering fine-grained face and hand mesh recovery since the kinematics tree is much deeper and the distal joints (e.g., fingers and face) are further away from the root joint.

\begin{figure*}[t]
    \begin{center}
        \includegraphics[width=\linewidth]{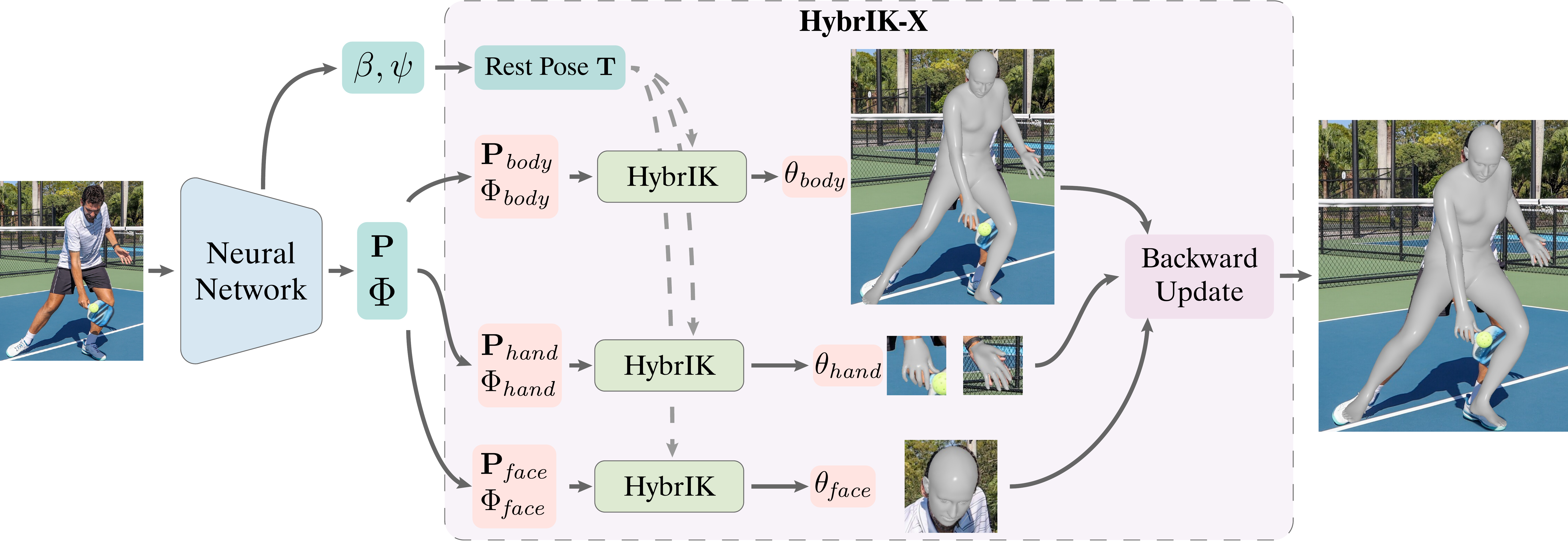}
    \end{center}
    \caption{\textbf{Illustration of the whole-body reconstruction pipeline.}
    The whole-body 3D joints $\mathbf{P}$, \textit{twist} angles $\Phi$, shape parameters $\beta$, and expression parameters $\psi$ are regressed from a holistic model. These results are split to three kinematic sub-trees and employ HybrIK independently. We solve the conflict joint positions and merge the solved poses from each sub-tree with the proposed backward update technique.
    }
    \label{fig:hybrikx}
\end{figure*}

~

\noindent
\textbf{HybrIK-X.}
To achieve well-aligned whole-body mesh recovery, we further present HybrIK-X. As depicted in Fig.~\ref{fig:hybrikx}, the core of HybrIK-X is a divide-and-conquer IK process. Specifically, we first divide the whole-body kinematic tree into four sub-trees (namely the left/right hands, face, and body), which have shorter lengths compared to the whole-body tree. By doing so, the distal joints in each sub-tree are closer to their corresponding root joint. Thus the sub-trees are more robust to noisy bone lengths and have a better model-image alignment. Subsequently, we apply HybrIK independently in each sub-tree. The recovered mesh of each sub-tree will align with its corresponding root joint ($\mathtt{left/right~wrists}$, $\mathtt{head}$, and $\mathtt{pelvis}$). Finally, we utilize a novel backward-updated algorithm to merge the results of all sub-trees and resolve the conflict joints.

The conflict joints refer to those joints that appear in two sub-trees concurrently, such as the $\mathtt{left/right~wrists}$ and $\mathtt{head}$. These joints are considered both the root joints of the sub-trees and the distal joints of the body-tree. If we directly combine the solved rotations of 4 sub-trees, the positions of the distal joints will be only determined by the results from the body-tree because the blend skinning function of SMPL-X depends on the feedforward FK process.
In order to preserve the well-aligned results from the sub-trees, we propose a backward-updated algorithm for recalculating the rotations of the parents of the conflict joints.

Given the estimated position of the distal joint $p_k$ and the reconstructed position of its grandparent joint $q_{\mathtt{pa}^2(k)}$, our objective is to recalculate the rotation $R_{\mathtt{pa}(k)}^*$ that satisfies:
\begin{align}
    R_{\mathtt{pa}(k)}^* = &\mathop{\text{argmin}}_{R_{\mathtt{pa}(k)}} \| R_{\mathtt{pa}(k)}\vec{t}_{\mathtt{pa}(k)} + q_{\mathtt{pa}^2(k)} - p_{\mathtt{pa}(k)} \|^2, \label{eq:prob1}\\
    s.t. \quad &\exists R_k \in \mathcal{SO}(3), ~ p_k = R_k\vec{t}_k + R_{\mathtt{pa}(k)}^*\vec{t}_{\mathtt{pa}(k)} + q_{\mathtt{pa}^2(k)}. \nonumber
\end{align}
By recalculating the rotation of the parent joint, the distal joint position will be consistent with the root position of the sub-tree. However, Eq.~\ref{eq:prob1} involves the $\mathcal{SO}(3)$ space and is not easy to solve. To make this problem solvable in a differentiable manner, we reformulate Eq.~\ref{eq:prob1} to an equivalent problem:
\begin{equation}
    \begin{aligned}
    q_{\mathtt{pa}(k)}^* = &\mathop{\text{argmin}}_{q_{\mathtt{pa}(k)}} \| q_{\mathtt{pa}(k)} - p_{\mathtt{pa}(k)} \|^2, \\
    s.t. \quad
    &\left\{
    \begin{aligned}
    \| q_{\mathtt{pa}(k)}^* - q_{\mathtt{pa}^2(k)} \| &= \| \vec{t}_{\mathtt{pa}(k)} \|, \\
    \| p_k - q_{\mathtt{pa}(k)}^* \| &= \| \vec{t}_k \|.
    \end{aligned}
    \right.
    \end{aligned}
\end{equation}
The analytical solution of the position of the parent joint $q_{\mathtt{pa}(k)}^*$ can be easily solved and computed differentiably. Detailed derivations are provided in appendix \S\ref{sec:bu}. Once we obtain $q_{\mathtt{pa}(k)}^*$, we follow the proposed twist-and-swing decomposition to calculate $R_{\mathtt{pa}(k)}$ and $R_k$.

~

\noindent
\textbf{Robust Jaw Pose Estimation.} The $\mathtt{jaw}$ rotation in the SMPL-X model relies on two joints, namely $\mathtt{jaw}$ and $\mathtt{head}$. However, accurately determining the position of the $\mathtt{head}$ joint can be challenging, particularly when people are wearing hats or looking upwards. Therefore, using Eq.~\ref{eq:swing-axis} and Eq.~\ref{eq:swing-angle} to calculate the swing rotation can result in an anomalous mouth-opening posture. To address this issue, we calculate $\mathtt{jaw}$ rotation by measuring the angle of mouth opening. In particular, we replace the $\mathtt{jaw}$ node in the kinematic tree with two nodes picked from the SMPL-X mesh ($\mathtt{mouth~top}$ and $\mathtt{mouth~bottom}$). We then utilize these two joints to determine the swing axis and swing angle, and subsequently obtain the swing rotation for the $\mathtt{jaw}$ joint using Eq.\ref{eq:swing-rod}.

\subsection{Learning Framework}\label{sec:learning}
The overall framework of our approach is illustrated in Fig.~\ref{fig:pipeline}. Firstly, a neural network is utilized to predict 2.5D joints $\mathbf{P}_{2.5D}$, \textit{twist} angles $\Phi$, shape parameters $\beta$, expression parameters $\psi$, and the initial camera parameter $s^0$. The 2.5D joints $\mathbf{P}_{2.5D}$ and the initial camera parameter $s^0$ are sent to the iterative camera estimation module to obtain the final camera prediction and the 3D joints $\mathbf{P}$. Secondly, the shape and expression parameters are used to obtain the rest pose $\mathbf{T}$ from the SMPL/SMPL-X model. Then, by combining $\mathbf{P}$, $\mathbf{T}$ and $\Phi$, we employ HybrIK/HybrIK-X to solve the rotations $\mathbf{R}$ of the 3D body, i.e., the pose parameters $\theta$. Finally, with the function $\mathcal{M}(\theta, \beta, \psi)$ provided by the SMPL/SMPL-X model, the whole-body mesh $M$ is obtained. The reconstructed pose $\mathbf{Q}$ can be obtained from $M$ by FK or a regressor, which is guaranteed to be realistic. Since HybrIK and HybrIK-X are differentiable, the whole framework is trained in an end-to-end manner.

~

\noindent
\textbf{Regression-based 3D Keypoint Estimation.}
Previous approaches to 3D keypoint estimation use heatmaps to represent the likelihood of joint positions. However, this heatmap representation costs a significant computational burden. Besides, it limits the output range within the input bounding box, which fails in the truncated scenarios where part of the human body is outside the input image. In real-world challenging scenarios, object detection methods are not perfect and always generate truncated bounding boxes when people are occluded or partially visible.

To reduce the computational cost and improve the robustness to real-world challenging scenarios, we adopt RLE~\cite{li2021human}, a simple yet effective regression paradigm. The conventional RLE leverages a fully-connected layer to estimate the joint coordinates $p_k \in \mathbb{R}^3$ and the standard deviation of each coordinate value $\sigma_k \in \mathbb{R}^3$. However, due to the limited diversity of current 3D human pose datasets, predicting three-dimensional deviations makes the learned distribution overfit to the training set, leading to an unstable training process. Here, we propose to use one-dimensional deviations to reduce the parameters of the learnable distribution and prevent overfitting, i.e., $\sigma_k \in \mathbb{R}$. When calculating the negative log-likelihood loss, we assume the coordinates of the three axes share the same standard deviation $\sigma_k$. The loss to train the 3D keypoint is formulated as:
\begin{equation}
    \mathcal{L}_{\textit{pose}} = - \sum_{k=1}^K \log Q(\bar{p}_k) - \log G_{\psi}(\bar{p}_k) + 3\cdot\log {\sigma}_k,
\end{equation}
where $Q(\cdot)$ is the probability density function of the standard Laplace distribution, $G_\psi$  is the distribution learned by RLE~\cite{li2021human}, and $\bar{p}_k = (p_k - \hat{p}_k) / \sigma_k$ and $\hat{p}_k$ denotes the ground-truth joint position. In practice, the network output for each joint is a 2.5D coordinate, i.e., the two-dimensional pixel coordinate with a relative depth value. To obtain the 3D coordinates, we back-project the 2.5D point to the 3D space with the estimated camera parameters.

With the regression paradigm, we get rid of the heatmap representation, and the model can be trained to infer the invisible body joints in challenging scenarios. Besides, the predicted deviation $\sigma_k$ can serve as the uncertainty index to evaluate the reliability of predicted keypoints, which is essential for downstream applications.

~

\noindent\textbf{Iterative Camera Estimation.} To back-project the 2.5D points onto the 3D space, we employ the weak-perspective camera model with a focal length of $1$ m. In contrast to previous work~\cite{hmr,spin,pixie}, we solely regress the scale factor $s \in \mathbb{R}$ and discard the translation, since we can determine the 2D position of the person by the root joint position. However, inferring the scale factor $s$ from a monocular RGB image is an ill-posed problem. To address this, we propose an iterative camera estimation method (ICE) that fully exploits the 2D visual cues, human body structure, and the power of the deep neural network for accurate and stable camera estimation. Specifically, given the currently estimated scale factor $s^t$ in the $t$-th step, we back-project the 2.5D points to 3D points:
\begin{equation}
    \mathbf{P}^t = \{p^t_k\}^K_{k=1} = \Pi^{-1}(\mathbf{P}_{2.5D}; s^t),
\end{equation}
where $\Pi(\cdot)$ is the projection function and $\Pi^{-1}(\cdot)$ denotes the back-projection function. Since the initial scale prediction is not correct, the projected 3D joints $\mathbf{P}^t$ might be wrong, e.g., overly small or overly large. We then input $\mathbf{P}^t$ to HybrIK-X to reconstruct the whole-body pose and retrieve reconstructed keypoints $\mathbf{Q}^t$, which satisfy the constraints imposed by the human body structure. We then use the least square method to analytically calculate the updated $s^{t+1}$ that minimizes projection error between $\mathbf{Q}^t$ and the detected 2D keypoints:
\begin{equation}
    s^{t+1} = \arg\min_{s^*} \| \Pi( \textit{IK}(\Pi^{-1}(\mathbf{P}_{2.5D}; s^{t})) ; s^*) - \mathbf{P}_{2D} \|^2.
\end{equation}
In this way, the human body structure and the 2D visual cues can provide a more accurate scale estimation. In the next iteration, a more accurate scale $s^{t+1}$ can obtain more accurate back-projected keypoints $\mathbf{P}^{t+1}$ and reconstructed keypoints $\mathbf{Q}^{t+1}$. Such an iterative update can alleviate the ambiguity in camera parameter estimation and generate stable results. The initially estimated $s^0$ is regressed by the neural network and supervised by the $\ell 2$ loss:
\begin{equation}
    \mathcal{L}_{\textit{cam}} = \| s^0 - \hat{s} \|^2,
\end{equation}
where $\hat{s}$ denotes the ground-truth scale factor.

~

\noindent
\textbf{Twist Angle Estimation.}
Instead of directly regressing the scalar value $\phi_k$, we choose to learn a 2-dimensional vector $(\cos{\phi_k}, \sin{\phi_k})$ to avoid the discontinuity problem. The $\ell 2$ loss is applied:
\begin{equation}
    \mathcal{L}_{\mathit{tw}} = \frac{1}{K} \sum_{k=1}^K \| (\cos{\phi_k}, \sin{\phi_k}) -  (\cos{\hat{\phi}_k}, \sin{\hat{\phi}_k}) \|^2,
\end{equation}
where $\hat{\phi}_k$ denotes the ground-truth \textit{twist} angle for the $k$-th joint.

~

\noindent
\textbf{Collaboration with SMPL-X.} The SMPL-X model allows us to obtain the rest pose skeleton with the additive offsets according to the shape parameters $\beta$ and expression parameters $\psi$:
\begin{equation}
    \mathbf{T} = W(\bar{M}_{\mathbf{T}} + B_{S}(\beta) + B_{E}(\psi)),
\end{equation}
where $\bar{M}_{\mathbf{T}}$ is the mesh vertices of mean rest pose, $B_{S}(\beta)$ and $B_{E}(\psi)$ are the blend shape functions provided by SMPL-X.
Then the pose parameters $\theta$ are calculated by HybrIK-X in a differentiable manner.
In the training phase, we supervise the shape parameters $\beta$:
\begin{equation}
    \mathcal{L}_{\textit{shape}} = \| \beta - \hat{\beta} \|^2,
\end{equation}
the expression parameters $\psi$:
\begin{equation}
    \mathcal{L}_{\textit{exp}} = \| \psi - \hat{\psi} \|^2,
\end{equation}
and the rotation parameters $\theta$:
\begin{equation}
    \mathcal{L}_{\textit{rot}} = \| \theta - \hat{\theta} \|^2.
\end{equation}
The overall loss of the learning framework is formulated as:
\begin{equation}
    \mathcal{L} = \mathcal{L}_{\textit{pose}} + \mu_1\mathcal{L}_{\textit{cam}} + {\mu_2}\mathcal{L}_{\textit{shape}} + {\mu_3}\mathcal{L}_{\textit{exp}} + {\mu_4}\mathcal{L}_{\textit{rot}} + {\mu_5}\mathcal{L}_{\textit{tw}},
\end{equation}
where $\mu_1, \mu_2$, $\mu_3$, $\mu_4$ and $\mu_5$ are weights of the loss items.

\subsection{Implementation Details}\label{sec:implement}

Here we elaborate more implementation details. We use HRNet-W48~\cite{sun2019deep} as the network backbone by default, initialized with ImageNet pre-trained weights. The HRNet output is fed to an average pooling layer, followed by the fully-connected layers to regress $\beta$, $\psi$, $\phi$, $\mathbf{P}_{2.5D}$ and $s^0$. The input image is resized to $256 \times 256$. The learning rate is set to $1\times10^{-3}$ at first and reduced by a factor of 10 at the 90th and 120th epoch. We use the Adam solver and train for 140 epochs, with a mini-batch size of $32$ per GPU and $4$ GPUs in total. In all experiments, $\mu_1 = \mu_2 = \mu_3 = 1$ and $\mu_4 = \mu_5 = 1\times10^{-2}$. Implementation is in PyTorch.

\begin{figure*}[!t]
    \begin{center}
        \includegraphics[width=\linewidth]{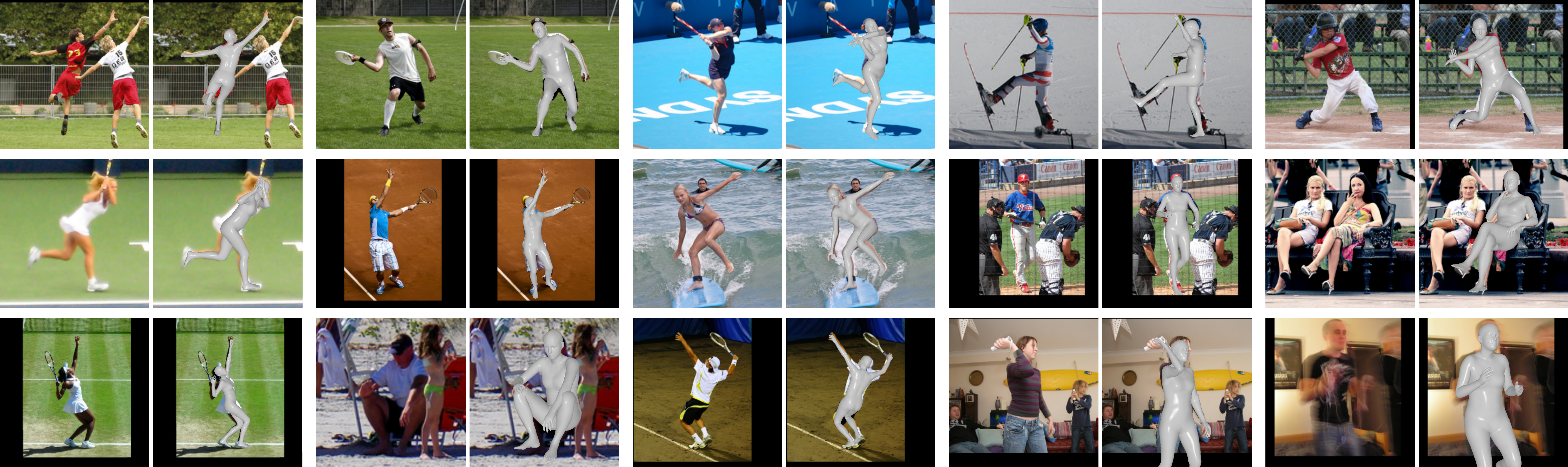}
    \end{center}
    \caption{\textbf{Qualitative results of body-only mesh recovery on the MSCOCO validation set} with challenging poses, occlusions, truncations, and motion blurs.
    }
    \label{fig:qualitative_body}
\end{figure*}

\begin{table*}[!th]
    \begin{center}
    \caption{\textbf{Benchmark of state-of-the-art models on 3DPW, Human3.6M and MPI-INF-3DHP datasets.} ``$*$'' denotes the method is trained on different datasets. ``-'' shows the results that are not available.}
    \label{tab:body}
    {%
        \begin{tabular}{l|ccc|cc|ccc}
        \toprule
        & \multicolumn{3}{c}{3DPW} & \multicolumn{2}{c}{Human3.6M} & \multicolumn{3}{c}{MPI-INF-3DHP} \\
        \cmidrule(lr){2-4} \cmidrule(lr){5-6} \cmidrule(lr){7-9}
        Method & PA-MPJPE $\downarrow$ & MPJPE $\downarrow$ & PVE $\downarrow$ & PA-MPJPE $\downarrow$ & MPJPE $\downarrow$ & PCK $\uparrow$ & AUC $\uparrow$ & MPJPE $\downarrow$ \\
        \midrule
        SMPLify~\cite{bogo2016keep} & - & - & - & 82.3 & - & - & - & - \\
        HMR~\cite{hmr} & 81.3 & 130.0 & - & 56.8 & 88.0 & 72.9 & 36.5 & 124.2 \\
        Pavlakos et al.~\cite{pavlakos2018learning} & - & - & - & 75.9 & - & - & - & - \\
        SPIN~\cite{spin} & 59.2 & 96.9 & 116.4 & 41.1 & - & 76.4 & 37.1 & 105.2 \\
        I2L-MeshNet~\cite{i2l}$^*$ & 58.6 & 93.2 & - & 41.7 & 55.7 & - & - & - \\
        KAMA~\cite{iqbal2021kama} & 51.1 & - & 97.0 & 40.2 & - & - & - & - \\
        ROMP~\cite{sun2021monocular} & 47.3 & 76.7 & 93.4 & - & - & - & - & - \\
        METRO~\cite{lin2021end} & 47.9 & 77.1 & 88.2 & 36.7 & 54.0 &- & - & - \\
        PARE~\cite{kocabas2021pare} & 46.5 & 74.5 & 88.6 & - & - & - & - & - \\
        \midrule
        HybrIK (ResNet-34) & {44.6} & {72.5} & {89.1} & 33.7 & 55.3 & 86.5 & 46.9 & 93.3 \\
        HybrIK (HRNet-W48) & \textbf{41.8} & \textbf{71.6} & \textbf{82.3} & \textbf{29.8} & \textbf{47.0} & \textbf{87.1} & \textbf{47.3} & \textbf{91.0} \\
        \bottomrule
        \end{tabular}
    }
    \end{center}
\end{table*}

\begin{table}[t]
    \begin{center}
    \caption{\textbf{Quantitative comparisons with state-of-the-art body-only methods on the AGORA dataset.}}
    \label{tab:agora}
    \resizebox{\linewidth}{!}
    {%
        \begin{tabular}{l|c|c|c|c}
        \toprule
        Method & NMVE $\downarrow$ & NMJE $\downarrow$ & MVE $\downarrow$ & MPJPE $\downarrow$ \\
        \midrule
        HMR~\cite{hmr} & 217.0 & 226.0 & 173.6 & 180.5 \\
        SPIN~\cite{spin} & 193.4 & 199.2 & 148.9 & 153.4 \\
        EFT~\cite{joo2021exemplar} & 196.3 & 203.6 & 159.0 & 165.4 \\
        PARE~\cite{kocabas2021pare} & 167.7 & 174.0 & 140.9 & 146.2 \\
        SPEC~\cite{kocabas2021spec} & 126.8 & 133.7 & 106.5 & 112.3 \\
        ROMP~\cite{sun2021monocular} & 113.6 & 118.8 & 103.4 & 108.1 \\
        PyMAF~\cite{zhang2021pymaf} & 200.2 & 207.4 & 168.2 & 174.2 \\
        BEV~\cite{sun2022putting} & 108.3 & 113.2 & 100.7 & 105.3 \\
        Hand4Whole~\cite{moon2022accurate} & 90.2 & 95.5 & 84.8 & 89.8 \\
        CLIFF~\cite{li2022cliff} & 83.5 & 89.0& 76.0 & 81.0 \\
        \midrule
        HybrIK & \textbf{81.2} & \textbf{84.6} & \textbf{73.9} & \textbf{77.0} \\
        \bottomrule
        \end{tabular}
    }
    \end{center}
\end{table}

\section{Empirical Evaluation}\label{sec:exp}

In this section, we first describe the datasets employed for training and quantitative evaluation. Next, we compare HybrIK and HybrIK with state-of-the-art approaches on body-only, hand-only, and whole-body mesh recovery benchmarks. Finally, ablation experiments are conducted to evaluate HybrIK and HybrIK-X.

\subsection{Datasets}
Following previous work, we train the body-only HybrIK on the Human3.6M~\cite{h36m}, 3DPW~\cite{3dpw}, MPI-INF-3DHP~\cite{3dhp}, MSCOCO~\cite{mscoco}, and AGORA~\cite{patel2021agora} datasets. For hand-only HybrIK, we train and evaluate on FreiHAND~\cite{zimmermann2019freihand} and HO3D-v2~\cite{hampali2020honnotate} datasets. For whole-body HybrIK-X, we use Human3.6M~\cite{h36m}, 3DPW~\cite{3dpw}, MPI-INF-3DHP~\cite{3dhp}, MSCOCO~\cite{mscoco}, and AGORA~\cite{patel2021agora} datasets for training. Detailed descriptions of the datasets are provided appendix \S\ref{sec:dataset}.

\begin{table*}[t]
    \begin{center}
    \caption{\textbf{Quantitative comparisons with state-of-the-art whole-body methods on the AGORA dataset.}}
    \label{tab:agorax}
    {%
        \begin{tabular}{l|cc|cc|cccc|cccc}
        \toprule
         & \multicolumn{2}{c}{{NMVE} $\downarrow$} & \multicolumn{2}{c}{{NMJE} $\downarrow$} & \multicolumn{4}{c}{{MVE} $\downarrow$} & \multicolumn{4}{c}{{MPJPE} $\downarrow$} \\
        \cmidrule(lr){2-13}
        Method & FB & B & FB & B & FB & B & F & LH/RH & FB & B & F & LH/RH \\
        \midrule
        SMPLify-X~\cite{pavlakos2019expressive} & 333.1 & 263.3 & 326.5 & 256.5 & 236.5 & 187.0 & 48.9 & 48.3/51.4 & 231.8 & 182.1 & 52.9 & 46.5/49.6 \\
        ExPose~\cite{choutas2020monocular} & 265.0 & 184.8 & 263.3 & 183.4 & 217.3 & 151.5 & 51.1 & 74.9/71.3 & 215.9 & 150.4 & 55.2 & 72.5/68.8 \\
        FrankMocap~\cite{rong2021frankmocap} & - & 207.8 & - & 204.0 & - & 168.3 & - & 54.7/55.7 & - & 165.2 & - & 52.3/53.1 \\
        PIXIE~\cite{pixie} & 233.9 & 173.4 & 230.9 & 171.1 & 191.8 & 142.2 & 50.2 & 49.5/49.0 & 189.3 & 140.3 & 54.5 & 46.4/46.0 \\
        Hand4Whole~\cite{moon2022accurate} & 144.1 & 96.0 & 141.1 & 92.7 &135.5 & 90.2 & 41.6 & 46.3/48.1 & 132.6 & 87.1 & 46.1 & 44.3/46.2 \\
        PyMAF-X~\cite{zhang2022pymaf} & 141.2 & 94.4 & 140.0 & 93.5 & 125.7 & 84.0 & \textbf{35.0} & 44.6/45.6 & 124.6 & 83.2 & \textbf{37.9} & 42.5/43.7 \\
        \midrule
        HybrIK-X & \textbf{120.5} & \textbf{73.7} & \textbf{115.7} & \textbf{72.3} & \textbf{112.1} & \textbf{68.5} & 37.0 & 46.7/47.0 & \textbf{107.6} & \textbf{67.2} & {38.5} & \textbf{41.2/41.4} \\
        \bottomrule
        \end{tabular}
    }
    \end{center}
\end{table*}

\subsection{Evaluation on Body-only Mesh Recovery}
To make a fair comparison with previous body-only mesh recovery methods, we use a regressor to obtain the $14$ LSP joints from the body mesh for the evaluation on 3DPW and Human3.6M datasets and $17$ joints for the MPI-INF-3DHP dataset. Procrustes aligned mean per joint position error (PA-MPJPE), mean per joint position error (MPJPE), percentage of correct keypoints (PCK), and area under curve (AUC) are reported to evaluate the 3D pose results. Per/Mean vertex error (PVE/MVE) is reported to evaluate the entire estimated body mesh. We further conduct experiments on the official AGORA test set. Normalized mean joint error (NMJE) and normalized mean vertex error (NMVE) are additionally reported.

In Tab.~\ref{tab:body}, we compare our method with previous 3D human pose and shape estimation methods, including both model-based and model-free methods, on 3DPW, Human3.6M, and MPI-INF-3DHP datasets.
Without bells and whistles, our method surpasses all previous state-of-the-art methods by a large margin on all three datasets. It is worth noting that our method improves $4.7$ mm PA-MPJPE (\textbf{10.1}\% relative improvement) on the 3DPW dataset, which shows that it is accurate and reliable to recover body mesh through inverse kinematics.

In Tab.~\ref{tab:agora}, we futher compare our method on the SMPL track of the AGORA test set. Compared to other 3D pose datasets, AGORA contains more challenging scenarios with severe occlusions and truncations. HybrIK shows consistent improvements on in this dataset. Qualitative results are shown in Fig.~\ref{fig:qualitative_body}.

\begin{table}[t]
    \begin{center}
    \caption{\textbf{Quantitative comparisons with state-of-the-art methods on the FreiHAND dataset.}}
    \label{tab:freihand}
    \resizebox{\linewidth}{!}
    {%
        \begin{tabular}{ll|c|c|c}
        \toprule
        & Method & PA-PVE $\downarrow$ & PA-MPJPE $\downarrow$ & F-Score @5mm $\uparrow$ \\
        \midrule
        \parbox[t]{1mm}{\multirow{6}{*}{\rotatebox[origin=c]{90}{Hand-only}}}
        & FreiHAND~\cite{zimmermann2019freihand} & 10.7 & - & 0.529 \\
        & Hasson et al.~\cite{hasson2019learning} & 13.2 & - & 0.436 \\
        & Boukhayma et al.~\cite{boukhayma20193d} & 13.0 & - & 0.435 \\
        & Pose2Mesh~\cite{choi2020pose2mesh} & 7.8 & 7.7 & 0.674 \\
        & I2L-MeshNet~\cite{i2l} & 7.6 & 7.4 & 0.681 \\
        & METRO~\cite{lin2021end} & 6.3 & 6.5 & 0.731 \\
        \midrule
        \parbox[t]{1mm}{\multirow{6}{*}{\rotatebox[origin=c]{90}{Whole-body}}}
        & ExPose~\cite{choutas2020monocular} & 11.8 & 12.2 & 0.484 \\
        & Zhou et al.~\cite{zhou2021monocular} & - & 15.7 & - \\
        & FrankMocap~\cite{rong2021frankmocap} & 11.6 & 9.2 & 0.553 \\
        & PIXIE~\cite{pixie} & 12.1 & 12.0 & 0.468 \\
        & Hand4Whole~\cite{moon2022accurate} & 7.7 & 7.7 & 0.664 \\
        & PyMAF~\cite{zhang2022pymaf} & 8.1 & 8.4 & 0.638 \\
        \midrule
        & HybrIK & \textbf{6.2} & \textbf{6.0} & \textbf{0.761} \\
        \bottomrule
        \end{tabular}
    }
    \end{center}
\end{table}

\begin{table}[t]
    \begin{center}
    \caption{\textbf{Quantitative comparisons with state-of-the-art methods on the HO-3D dataset.}}
    \label{tab:ho3d}
    \resizebox{\linewidth}{!}
    {%
        \begin{tabular}{l|c|c|c}
        \toprule
        Method & PA-PVE $\downarrow$ & PA-MPJPE $\downarrow$ & F-Score @5mm $\uparrow$ \\
        \midrule
        Pose2Mesh~\cite{choi2020pose2mesh} & 1.27 & 1.25 & 0.441 \\
        I2L-MeshNet~\cite{i2l} & 1.39 & 1.12 & 0.409 \\
        I2UV-HandNet~\cite{chen2021i2uv} & 1.01 & \textbf{0.99} & 0.500 \\
        METRO~\cite{lin2021end} & 1.11 & 1.04 & 0.484 \\
        Hampali et al.~\cite{hampali2020honnotate} & 1.06 & 1.07 & 0.506 \\ 
        Hasson et al.~\cite{hasson2019learning} & 1.12 & 1.10 & 0.464 \\
        Hasson et al.~\cite{hasson2020leveraging} & 1.14 & 1.14 & 0.428\\
        ArtiBoost~\cite{li2021artiboost} & 1.09 & 1.14 & 0.488 \\
        Keypoint Trans.~\cite{hampali2022keypoint} & - & 1.08 & - \\
        \midrule
        HybrIK & \textbf{0.96} & \textbf{0.99} & \textbf{0.550} \\
        \bottomrule
        \end{tabular}
    }
    \end{center}
\end{table}

\subsection{Evaluation on Hand-only Mesh Recovery}
To validate the generalization of our inverse kinematics solution, we conduct experiments on two widely-used hand pose benchmarks: FreiHAND~\cite{zimmermann2019freihand} and HO3D~\cite{hampali2020honnotate}. MPJPE and PVE are reported for evaluation. F-Score~\cite{knapitsch2017tanks} is also reported to evaluate the harmonic mean between the predicted vertices and the ground-truth vertices.

The comparisons of HybrIK with previous state-of-the-art methods are reported in Tab.~\ref{tab:freihand} and Tab.~\ref{tab:ho3d}. It is noteworthy that, as observed in prior research~\cite{i2l,zhang2022pymaf}, recent hand-only methods~\cite{i2l,lin2021end} that adopt non-parametric representation exhibit a numerical advantage over parametric-based methods. HybrIK significantly outperforms the most accurate whole-body method by 1.5 mm MPJPE (\textbf{19.5}\% relative improvement) on the FreiHAND dataset. Additionally, even when compared against the state-of-the-art hand-only methods, HybrIK exhibits a \textbf{7.5}\% relative improvement. Compared to the methods designed specifically for challenging interaction scenarios on the HO3D dataset, HybrIK also exhibits state-of-the-art performance.

\begin{figure}[t]
    \begin{center}
        \includegraphics[width=\linewidth]{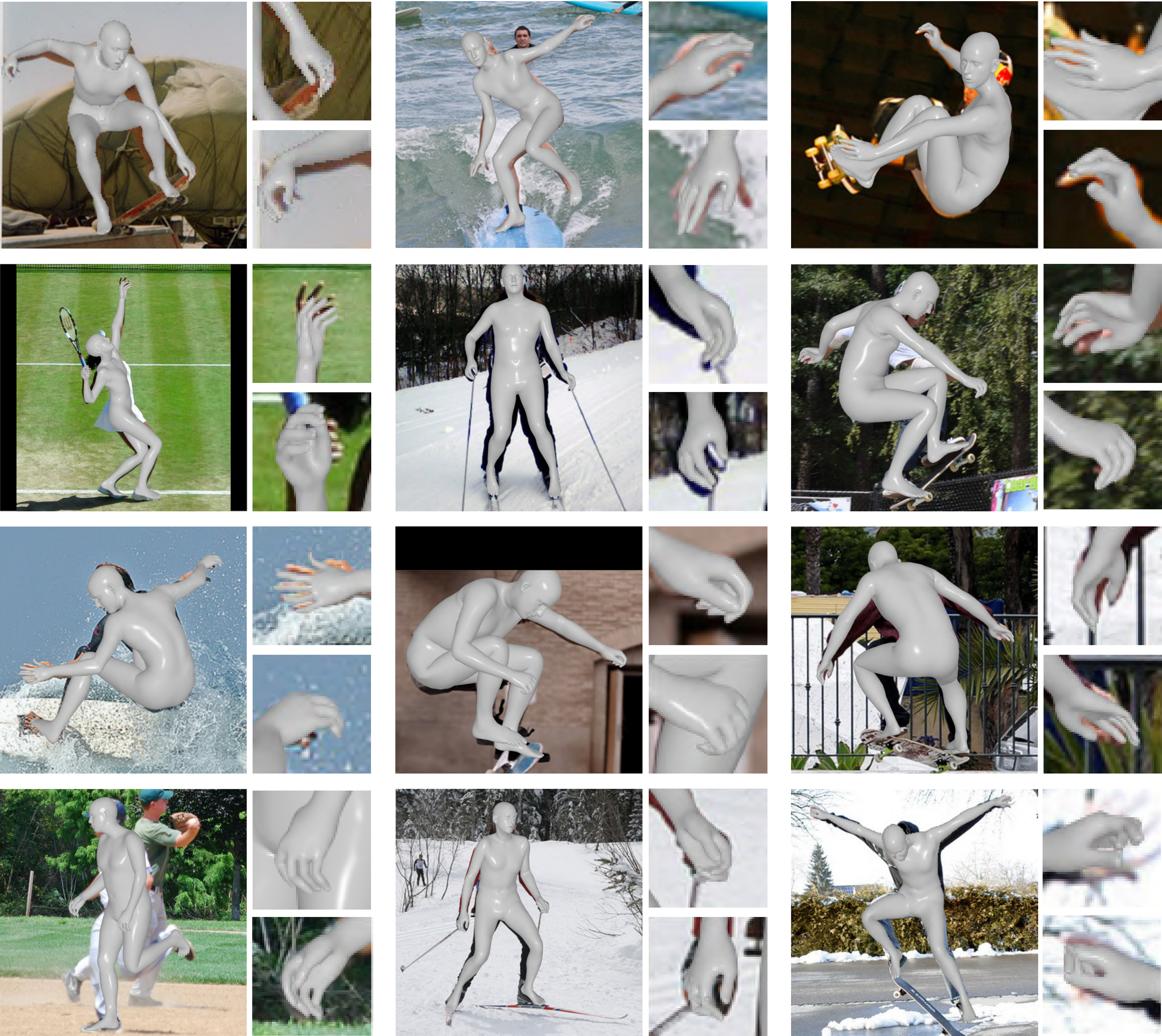}
    \end{center}
    \caption{\textbf{Qualitative results of whole-body mesh recovery on the MSCOCO validation set.}
    }
    \label{fig:qualitative}
\end{figure}

\subsection{Evaluation on Whole-body Mesh Recovery}

To evaluate our hybrid inverse kinematics solution on whole-body mesh recovery, we conduct experiments on the SMPL-X track of the AGORA test set. The evaluation on AGORA is affected by the detection results since there are multiple persons on one image. We follow previous work~\cite{moon2022accurate,zhang2022pymaf} and use the same detection results for a fair comparison.

The evaluation is depicted in Tab.~\ref{tab:agorax}. Note that previous methods use separate expert networks to handle body, face, and hand estimation. They cost higher computational resources and longer running time. HybrIK-X is a one-stage approach and demonstrates a significant superiority over the state-of-the-art whole-body methods, achieving \textbf{24.3} mm and \textbf{20.7} mm improvement in full-body NMJE and NVME, respectively. For the fine-grained face and hand results, HybrIK-X obtains comparable performance against the most accurate method with a much smaller input resolution and less computation. Qualitative results on the MSCOCO validation set are shown in Fig.~\ref{fig:qualitative}. Qualitative comparison with state-of-the-art approaches are provided in appendix \S\ref{sec:qualitative}. Detailed comparisons of computation complexity are reported in \S\ref{sec:ablation}.

\subsection{Ablation Study}\label{sec:ablation}

In this study, we evaluate the effectiveness of the twist-and-swing decomposition and the proposed inverse kinematics algorithms. Evaluation is conducted on the AGORA validation set by default as it contains challenging in-the-wild scenarios. More experimental results are provided in appendix \S\ref{sec:supp_ablation}.

\begin{figure}[t]
    \begin{center}
        \includegraphics[width=\linewidth]{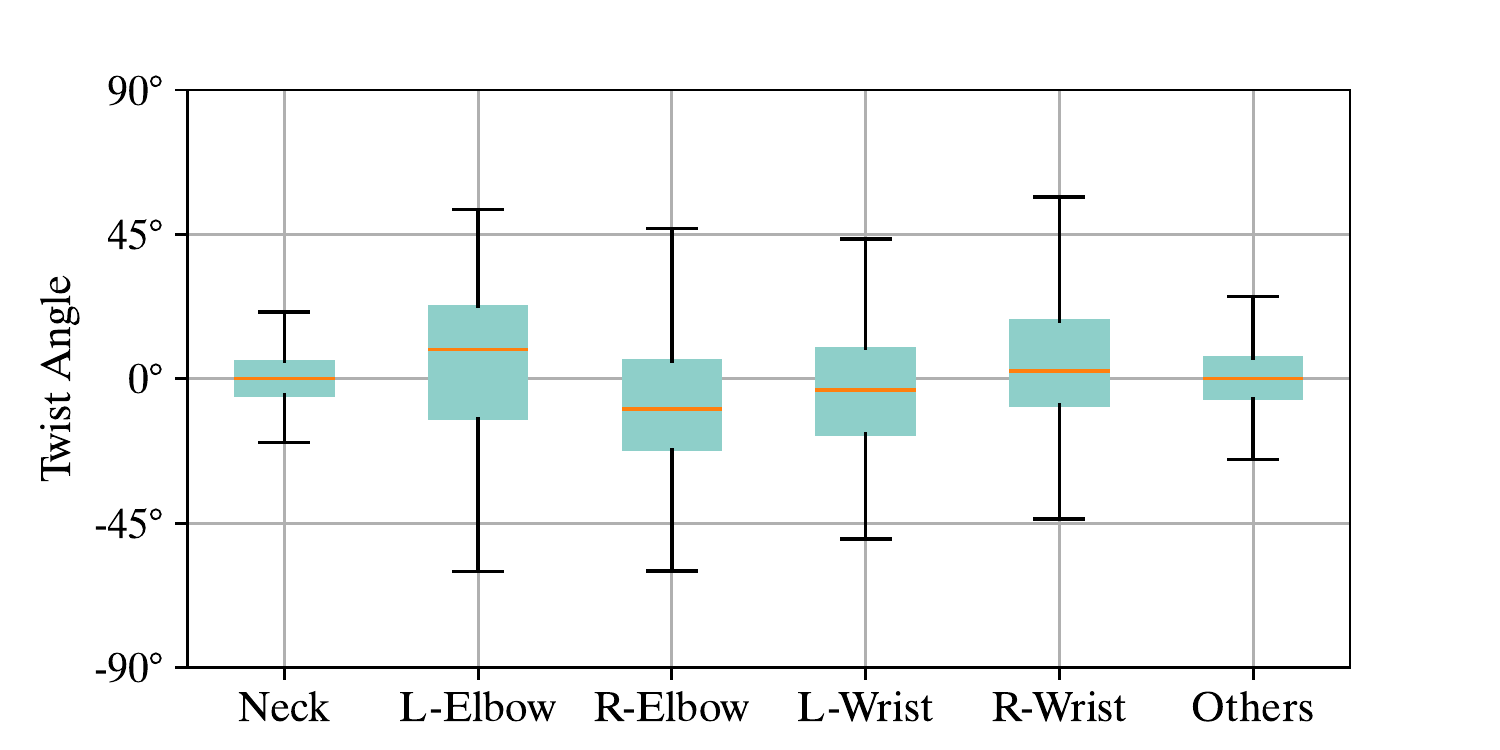}
    \end{center}
    \caption{\textbf{Distribution of \textit{twist} angles}. Only a few joints have a range over $30^{\circ}$. Other joints have a limited range of \textit{twist} angle.}
    \label{fig:dist-twist}
\end{figure}

\begin{table}[t]
    \begin{center}
    \caption{\textbf{Reconstruction error with different \textit{twist} angle.} The accurate \textit{twist} angles significantly reduce the reconstruction error.}
    \label{tab:twist}
    \resizebox{\linewidth}{!}
    {%
        \setlength{\tabcolsep}{1mm}{
        \begin{tabular}{l|ccc|ccc|ccc}
        \toprule
        & \multicolumn{3}{c}{Random Twist} & \multicolumn{3}{c}{Estimated Twist} & \multicolumn{3}{c}{Zero Twist} \\
        \cmidrule(lr){2-4} \cmidrule(lr){5-7} \cmidrule(lr){8-10}
        ~& 24 \textit{jts} & 14 \textit{jts} & Vert. & 24 \textit{jts} & 14 \textit{jts} & Vert. & 24 \textit{jts} & 14 \textit{jts} & Vert.  \\
        \midrule
        Error & 0.1 & 40.0 & 67.3 & 0.1 & 6.1 & 10.0 & 0.1 & 6.8 & 12.1 \\
        \bottomrule
        \end{tabular}
        }
    }
    \end{center}
\end{table}

\begin{table}[t]
    \begin{center}
    \caption{\textbf{Robustness to noisy joints}. Mean errors of body and hand joints are reported.}
    \label{tab:robust}
    \resizebox{\linewidth}{!}
    {
        \begin{tabular}{l|c|c|c|c}
            \toprule
            
            & \multicolumn{4}{c}{Body / Hand MPJPE} \\
            \cmidrule(lr){2-5}
            ~& GT Joints & $\pm 10$ mm & $\pm 20$ mm & $\pm 30$ mm \\
            \midrule
            Naive HybrIK & 0.1/0.1 & 44.4/57.6 & 74.5/89.5 & 104.7/121.9 \\
            Adaptive HybrIK & 0.1/0.1 & 21.6/16.5 & 40.6/32.3 & 59.0/47.8 \\
            HybrIK-X & 0.1/0.1 & \textbf{21.3/13.6} & \textbf{39.6/26.9} & \textbf{57.6/40.4} \\
            \bottomrule
        \end{tabular}
    }
    \end{center}
\end{table}

~

\noindent
\textbf{Analysis of the twist rotation.} To demonstrate the effectiveness of twist-and-swing decomposition, we first count the distribution of the \textit{twist} angle in the AGORA validation set. The distribution is illustrated in Fig.~\ref{fig:dist-twist}. As expected, due to the physical limitation, only $\mathtt{neck}$, $\mathtt{elbow}$ and $\mathtt{wrist}$ have a wide range of variations. All other joints have a limited range of \textit{twist} angle (less than 30$^\circ$).
It indicates that the \textit{twist} angle can be reliably estimated.

Besides, we develop an experiment to see how the \textit{twist} angles affect the reconstructed pose and shape. We take the ground-truth 24 SMPL joints and shape parameters as the input of the HybrIK process. As for the \textit{twist} angle, we compare random values in $[-{\pi}, {\pi}]$ and the values estimated by the network. We evaluation the mean error of the reconstructed 24 SMPL joints, the 14 LSP joints, the body mesh and the \textit{twist} angle. Here, following previous work~\cite{bogo2016keep,hmr,spin}, the 14 LSP joints are regressed from the body mesh by a pretrained regressor. Quantitative results are reported in Tab.~\ref{tab:twist}. It shows that the regressed \textit{twist} angles significantly reduce the error on the mesh vertices and the LSP joints that regressed from the mesh. Since most of the twist angles are close to zeros, the zero twist angles produce acceptable performance. Notice that the wrong \textit{twist} angles do not affect the reconstructed SMPL joints. Only the \textit{swing} rotations change the joint locations.

\begin{table}[t]
    \begin{center}
    \caption{\textbf{The error correction capability}.}
    \label{tab:pat-ik}
    {%
        \begin{tabular}{l|c|c|c}
        \toprule
        ~& Predicted Pose & HybrIK & SMPLify~\cite{bogo2016keep} \\
        \midrule
        MPJPE (24 \textit{jts}) $\downarrow$ & 73.1 mm & \textbf{66.3} mm & 100.1 mm \\
        \bottomrule
        \end{tabular}
    }
    \end{center}
\end{table}

\begin{table}[t]
    \begin{center}
    \caption{\textbf{Ablation experiments on the truncated AGORA validation set.} ``RLE$^{-}$'' denotes our improved version of RLE with reduced distribution parameters.}
    \label{tab:trunc}
    \resizebox{\linewidth}{!}
    {%
        \begin{tabular}{l|cc|cc}
        \toprule
        & \multicolumn{2}{c}{Validation w. Trunc.} & \multicolumn{2}{c}{Test} \\
        \cmidrule(lr){2-5}
        Method & MVE $\downarrow$ & MPJPE $\downarrow$ & MVE $\downarrow$ & MPJPE $\downarrow$ \\
        \midrule
        HybrIK-X (Heatmap-based) & 100.4 & 99.8 & 134.1 & 127.5 \\
        HybrIK-X (RLE) & {91.9} & {82.7} & 113.9 & 109.7 \\
        HybrIK-X (RLE$^-$) & \textbf{89.9} & \textbf{80.4} & \textbf{112.1} & \textbf{107.6} \\
        \bottomrule
        \end{tabular}
    }
    \end{center}
\end{table}

~

\noindent
\textbf{Robustness to Noisy Joint Positions.}
To demonstrate the superiority of HybrIK-X over Adaptive and Naive HybrIK, we compare the reconstruction errors with inputs in different noise levels on the AGORA validation set. We add jitters to the ground-truth joint positions and then feed them to the IK algorithms. Quantitative comparisons on whole-body recovery are reported in Tab.~\ref{tab:robust}. We report the reconstruction errors of body and hand joints separately. It shows that when the input joints are correct, all three algorithms introduce negligible errors. As the noise level increases, HybrIK-X and Adaptive HybrIK are more robust than Naive HybrIK. For body reconstruction, HybrIK-X has similar performance to Adaptive HybrIK. For fine-grained hand reconstruction, HybrIK-X exhibits better robustness against the noisy joint positions.

~

\noindent
\textbf{Error correction capability.}
In this experiment, we examine the error correction capability of the HybrIK algorithm. The HybrIK algorithm is fed with the 3D joints, \textit{twist} angles and shape parameters that predicted by the neural network. Additionally, we apply the SMPLify~\cite{bogo2016keep} algorithm on the predicted pose and compare it to our method. As shown in Tab.~\ref{tab:pat-ik}, the error of reconstructed joints after HybrIK is reduced the error to $66.3$mm, while SMPLify raises the error to 100.1 mm. The error correction capability of HybrIK comes from the fact that the network may predict unrealistic body pose, e.g., left-right asymmetry and abnormal limbs proportions. In contrast, the rest pose is generated by the parametric statistical body model, which guarantees that the reconstructed pose is consistent with the realistic body shape distribution. Since our proposed framework is agnostic to the way we obtain 3D joints, we can improve the performance of any 3D keypoint estimation approaches.

~

\noindent
\textbf{Effectiveness of the Regression Paradigm.}
We further evaluate the effectiveness of the regression paradigm. Since the AGORA test set does not provide ground-truth bounding boxes, the model performance will be affected by the object detection results. Besides, the test set is quite challenging because the persons in it are severely occluded and truncated. Therefore, the AGORA test set can serve as the benchmark to evaluate the robustness of the mesh recovery algorithms. Moreover, we augment the current validation set with challenging truncated bounding boxes. We simulate truncations by cropping each frame with a truncated window. The areas of the cropped windows are $3/4$ of the original ones. Quantitative comparisons about different 3D keypoint estimation paradigms are reported in Tab.~\ref{tab:trunc}. Heatmap cannot represent the joints outside the input bounding box. Therefore, there is a significant performance degradation when using heatmaps to obtain joint positions. Additionally, the improved version of RLE shows better performance than the original RLE since it introduces fewer parameters to the distribution and avoids overfitting. It is demonstrated that our method is robust to challenging applications with occlusions and truncations. Qualitative comparisons between the heatmap-based paradigm and the regression-based paradigm are provided in appendix \S\ref{sec:qualitative}.

\begin{table}[!t]
    \begin{center}
    \caption{\textbf{Results of depth estimation on the AGORA validation set}.}
    \label{tab:depth}
    {%
        \begin{tabular}{l|c|c|c|c|c|c}
        \toprule
        & \multicolumn{6}{c}{ICE Steps} \\
        \cmidrule(lr){2-7}
        ~ & 0 & 1 & 2 & 3 & 4 & 5 \\
        \midrule
        Error $\downarrow$ & 183.91 & 181.10 & 179.51 & 178.60 & 178.18  & 178.09 \\
        \bottomrule
        \end{tabular}
    }
    \end{center}
\end{table}

~

\noindent
\textbf{Effectiveness of Iterative Camera Estimation.}
To study the effectiveness of the proposed iterative camera estimation (ICE), we evaluate the performance with different iterative steps. We calculate the mean error of the estimated distance between the human body and the camera. Quantitative results are summarized in Tab.~\ref{tab:depth}. ``0 step'' is equivalent to direct regression without the iterative update. We can observe that ICE can improve the accuracy of the camera distance estimation. In practice, 3 steps are accurate enough.


\begin{table}[t]
    \begin{center}
    \caption{\textbf{Computation complexity and model parameters.} Full-body mean vertex error (FB-MVE) is also listed.}
    \label{tab:complex}
    {%
        \begin{tabular}{l|cc|c}
        \toprule
        Method & FLOPS $\downarrow$ & \#Params $\downarrow$  & FB-MVE\\
        \midrule
        PIXIE~\cite{pixie} & 31.0G & 192.9M & 191.8 \\
        PyMAF-X~\cite{zhang2022pymaf} & 46.5G & 203.6M & 125.7 \\
        \midrule
        HybrIK-X (One-Stage) & \textbf{22.7G} & \textbf{76.1M} & \textbf{112.1} \\
        \bottomrule
        \end{tabular}
    }
    \end{center}
\end{table}

~

\noindent
\textbf{Computation Complexity.} The experimental results of computation complexity and model parameters are listed in Tab.~\ref{tab:complex}. The proposed one-stage HybrIK-X achieves more accurate whole-body mesh recovery results with significantly lower computation complexity and fewer model parameters. Specifically, the total FLOPs are reduced by \textbf{26.8}\%, and the parameters are reduced by \textbf{60.5}\%. The efficiency and effectiveness of HybrIK-X are of great value in downstream applications.

\section{Conclusion}
In this paper, we present a hybrid analytical-neural inverse kinematics framework, HybrIK, for body-only mesh recovery. This framework is further extended to whole-body mesh recovery and named HybrIK-X. HybrIK and HybrIK-X transform the 3D joint locations to a pixel-aligned accurate human body mesh via inverse kinematics, and then obtains a more accurate and realistic 3D skeleton from the reconstructed 3D mesh with forward kinematics, closing the loop between the 3D skeleton and the parametric body model. To evaluate the effectiveness of our approach, we perform experiments on various benchmarks for body-only, hand-only, and whole-body scenarios. The results indicate that our approach outperforms existing state-of-the-art approaches by a significant margin. Moreover, the proposed approach is fully differentiable and uses a one-stage network to recover the whole-body mesh, making it considerably more efficient than existing approaches. Overall, we believe our approach can serve as a strong baseline for future research and offers a new perspective on whole-body mesh recovery.

\begin{appendices}

\section{Rigid Registration of Global Rotation}
\label{sec:svd}

In the SMPL/SMPL-X model~\cite{loper2015smpl,pavlakos2019expressive}, the pose parameters $\theta$ control the rotations of the rigid body parts. The three joints named $\mathtt{spine}$, $\mathtt{left~hip}$ and $\mathtt{right~hip}$ constitute a rigid body part, which is controlled by the global root rotation. Consequently, the global rotation can be ascertained by registering the rest pose template of $\mathtt{spine}$, $\mathtt{left~hip}$ and $\mathtt{right~hip}$ to the predicted locations of these three joints. Let $t_1$, $t_2$ and $t_3$ denote their locations in the rest pose template, and $p_1$, $p_2$ and $p_3$ denote the predicted locations. Our objective is to identify a rigid rotation that optimally aligns the two sets of joints. Here, we assume the root joint of the predicted pose and the rest pose are aligned. Hence, the problem is formulated as:
\begin{equation}
   R_0 = \arg\min_{R \in \mathbb{SO}^3} \sum_{i=1}^3 \| p_i - R t_i \|_2^2.
\end{equation}
This formula can be written in matrix form:
\begin{equation}
   R_0 = \arg\min_{R \in \mathbb{SO}^3} \| P_0 - R T_0 \|_F^2,
   \label{eq:problem}
\end{equation}
where $\| \cdot \|_F$ denotes the Frobenius norm,$P_0$ denotes $[p_0~p_1~p_2]$, and $T_0$ denotes $[t_0~t_1~t_2]$. Let us simplify the expression in Eq.~\ref{eq:problem} as:
\begin{equation}
   \begin{aligned}
      &\min_{R \in \mathbb{SO}^3} \| P_0 - R T_0 \|_F^2 \\
      \Leftrightarrow &\min_{R \in \mathbb{SO}^3} \text{trace}((P_0 - R T_0)^\mathrm{T}(P_0 - R T_0)) \\
      \Leftrightarrow &\min_{R \in \mathbb{SO}^3} \text{trace}(P_0^\mathrm{T}P_0 + T_0^\mathrm{T}T_0 - 2P_0^\mathrm{T}RT_0). \\
   \end{aligned}
\end{equation}
Note that $P_0^\mathrm{T}P_0$ and $T_0^\mathrm{T}T_0$ are independent of $R$. Thus the original problem is equivalent to:
\begin{equation}
   \begin{aligned}
   &\arg\min_{R \in \mathbb{SO}^3} \| P_0 - R T_0 \|_F^2 \\
   \Leftrightarrow &\arg\max_{R \in \mathbb{SO}^3} \text{trace}(P_0^\mathrm{T}RT_0).
   \end{aligned}
\end{equation}
Further, we can leverage the property of the matrix trace,\begin{equation}
   \text{trace}(P_0^\mathrm{T}RT_0) = \text{trace}(RT_0P_0^\mathrm{T}).
\end{equation}
Then, we apply Singular Value Decomposition (SVD) to the joint locations:
\begin{equation}
   T_0P_0^\mathrm{T} = U \Lambda V^\mathrm{T}.
\end{equation}
The problem is equivalent to:
\begin{equation}
   \begin{aligned}
   &\arg\max_{R \in \mathbb{SO}^3} \text{trace}(RT_0P_0^\mathrm{T}) \\
   \Leftrightarrow &\arg\max_{R \in \mathbb{SO}^3} \text{trace}(R U \Lambda V^\mathrm{T}) \\
   \Leftrightarrow &\arg\max_{R \in \mathbb{SO}^3} \text{trace}(\Lambda V^\mathrm{T}R U).
   \end{aligned}
\end{equation}
Note that $U$, $V$ and $R$ are orthogonal matrices, so $M = V^\mathrm{T}RU$ is also an orthogonal matrix. Then, for all $1 \leq j \leq$ we have:
\begin{equation}
   \begin{aligned}
   &m_j^\mathrm{T}m_j = 1 = \sum_{i=1}^3 m_{ij}^2 \\
   \Rightarrow ~ &m_{ij}^2 \leq 1 \Rightarrow | m_{ij} | \leq 1.
   \end{aligned}
\end{equation}
Besides, $\Lambda$ is a diagonal matrix
with non-negative values, i.e., $\lambda_1$, $\lambda_2$, $\lambda_3 \geq 0$. Therefore:
\begin{equation}
   \begin{aligned}
   \text{trace}(\Lambda V^\mathrm{T}R U) &= \text{trace}(\Lambda M) \\ &= \sum_{i=1}^3 \lambda_i m_{ii}
   \leq \sum_{i=1}^3 \lambda_i.
   \end{aligned}
\end{equation}
The trace is maximized if $m_{ii} = 1, \forall 1 \leq i \leq 3$. That means $M = \mathcal{I}$, where $\mathcal{I}$ is the identity matrix. Finally, the optimal rotation $R_0$ is:
\begin{equation}
   \begin{aligned}
   V^\mathrm{T}R_0 U = \mathcal{I} \\
   \Rightarrow
   R_0 = V U^\mathrm{T}.
   \end{aligned}
\end{equation}

\section{Backward-updated Algorithm}
\label{sec:bu}

In the backward-updated algorithm of HybrIK-X, our aim is to analytically calculate the position of the parent joint, denoted as $q_{\mathtt{pa}(k)}^*$. Recall that $q_{\mathtt{pa}(k)}^*$ should satisfy the following equation:
\begin{equation}
    \begin{aligned}
        q_{\mathtt{pa}(k)}^* = &\mathop{\text{argmin}}_{q_{\mathtt{pa}(k)}} \| q_{\mathtt{pa}(k)} - p_{\mathtt{pa}(k)} \|^2, \\
        s.t. \quad
        &\left\{
        \begin{aligned}
        \| q_{\mathtt{pa}(k)}^* - q_{\mathtt{pa}^2(k)} \| &= \| \vec{t}_{\mathtt{pa}(k)} \|, \\
        \| p_k - q_{\mathtt{pa}(k)}^* \| &= \| \vec{t}_k \|.
        \end{aligned}
        \right.
    \end{aligned}
\end{equation}
To simplify the notation, we define $A = q_{\mathtt{pa}^2(k)}$, $B = p_{\mathtt{pa}(k)} $, $B^* = q_{\mathtt{pa}(k)}^*$, and $C = p_k$ for the subsequent derivation. The aforementioned equation can be reformulated as:
\begin{align}
    B^* = &\mathop{\text{argmin}}_{q_{\mathtt{pa}(k)}} \| q_{\mathtt{pa}(k)} - B \|^2, \\
    s.t. \quad
    &\left\{
    \begin{aligned}
    \| B^* - A \| &= \| \vec{t}_{\mathtt{pa}(k)} \|, \\
    \| C - B^* \| &= \| \vec{t}_k \|.
    \end{aligned}\label{eq:norm}
    \right.
\end{align}
The vector $\overrightarrow{AB^*}$ can be orthogonally decomposed with respect to $\overrightarrow{AC}$ as follows:
\begin{equation}
    \overrightarrow{AB^*} = \vec{v}_{\parallel} + \vec{v}_{\perp},
\end{equation}
where $\vec{v}_{\parallel}$ is parallel to $\overrightarrow{AC}$ and $\vec{v}_{\perp}$ is perpendicular to $\overrightarrow{AC}$. We introduce a point $D$, such that $\overrightarrow{AD} = \vec{v}_{\parallel}$ and $\overrightarrow{DB^*} = \vec{v}_{\perp}$. Since $\vec{v}_{\parallel}$ is parallel to $\overrightarrow{AC}$, it can be expressed as $\overrightarrow{AD} = m \overrightarrow{AC}$, with $m \in [0, 1]$ representing the corresponding scalar. Consequently, by determining $k$ and $\vec{v}_{\perp}$, we can ascertain the position of $B^*$.

According to the norm constrains in Eq.~\ref{eq:norm}, we can determine $m$ as:
\begin{equation}
    \begin{gathered}
    \| \vec{t}_{\mathtt{pa}(k)} \|^2 - m^2\overrightarrow{AC}^2 = \| \vec{t}_k \|^2 - (1 - m)^2\overrightarrow{AC}^2, \\
    \Rightarrow m = \frac{\| \vec{t}_{\mathtt{pa}(k)} \|^2 - \| \vec{t}_k \|^2 + \overrightarrow{AC}^2}{2\overrightarrow{AC}^2}.
    \end{gathered}
\end{equation}
Once we determine $m$, the original problem can be written as:
\begin{equation}
    \overrightarrow{DB^*} = \mathop{\text{argmin}} \| \overrightarrow{DB} - \overrightarrow{DB^*} \|^2.
\end{equation}
$\overrightarrow{DB}$ can be orthogonally decomposed according to $\overrightarrow{AC}$ as:
\begin{equation}
    \overrightarrow{DB} = \overrightarrow{DB}_{\parallel} + \overrightarrow{DB}_{\perp},
\end{equation}
where $\overrightarrow{DB}_{\parallel} \parallel \overrightarrow{AC} \parallel \overrightarrow{DB^*} $ and $\overrightarrow{DB}_{\perp} \perp \overrightarrow{AC}$.

Thus, $\| \overrightarrow{DB} - \overrightarrow{DB^*} \|^2 = \| \overrightarrow{DB}_\perp - \overrightarrow{DB^*} \|^2 + \overrightarrow{DB}_\parallel^2$, where $\overrightarrow{DB}_\parallel$ is a constant an irrelevant to $B^*$. Intuitively, the optimal $\overrightarrow{DB^*}$ must be parallel to $\overrightarrow{DB}_\perp^2$ since they are both perpendicular to $\overrightarrow{AC}$ and sharing the same start point $D$. Consequently, we have $\overrightarrow{DB^*} = n \frac{\overrightarrow{DB}_\perp}{\| \overrightarrow{DB}_\perp \|}$, where $n$ denotes the norm of $\overrightarrow{DB^*}$. Following the norm constrains in Eq.~\ref{eq:norm}, we can determine $n$ as:
\begin{equation}
    n = \sqrt{\| \vec{t}_{\mathtt{pa}(k)} \|^2 - m^2\overrightarrow{AC}^2}.
\end{equation}

Finally, we can calculate $B^*$ as:
\begin{equation}
    \begin{aligned}
        B^* &= A + \overrightarrow{AB}, \\
        &= A + m\overrightarrow{AC} + n\frac{\overrightarrow{DB}_\perp}{\| \overrightarrow{DB}_\perp \|}, \\
    \end{aligned}
\end{equation}
where
\begin{equation}
    \overrightarrow{DB}_\perp = \overrightarrow{DB} - \frac{\overrightarrow{DB}\cdot\overrightarrow{AC}}{\| \overrightarrow{AC} \|^2} \overrightarrow{AC}.
\end{equation}

\section{Datasets}
\label{sec:dataset}

\noindent\textbf{AGORA~\cite{patel2021agora}:}\quad
It is a synthetic dataset featuring precise SMPL and SMPL-X annotations fitted to 3D scans. We use this dataset for training only when conducting experiments on it. The evaluation is performed on the official platform with separated SMPL and SMPL-X tracks.

~

\noindent\textbf{3DPW~\cite{3dpw}:}\quad
It is a challenging outdoor benchmark for body-only 3D pose and shape estimation. It contains $60$ video sequences obtained from a hand-held moving camera. It employs IMU sensors to calculate ground-truth pose and shape data.

~

\noindent\textbf{MPI-INF-3DHP~\cite{3dhp}:}\quad
It is a body-only 3D keypoint dataset with both constrained indoor and complex outdoor scenes. It includes 8 actors performing 8 activities from 14 camera views. Following \cite{hmr,spin}, we use its train set for training and evaluate on its test set.

~

\noindent\textbf{Human3.6M~\cite{h36m}:}\quad
It is an indoor benchmark for body-only 3D pose and shape estimation. It includes 11 subjects performing 15 different activities in a laboratory environment. Following \cite{hmr,spin}, we use 5 subjects (S1, S5, S6, S7, S8) for training and 2 subjects (S9, S11) for evaluation.

~

\noindent\textbf{FreiHAND~\cite{zimmermann2019freihand}:}\quad
It is a single-hand 3D pose dataset with MANO~\cite{romero2022embodied} annotations. It contains over 130k training samples with the right hands. We employ HybrIK on the hand pose model and use this dataset for evaluation.

~

\noindent\textbf{HO3D-v2~\cite{hampali2020honnotate}:}\quad
It is a dataset with 3D pose annotations for hands and objects under severe occlusions from each other. It contains sequences of the right hand interacting with an object. We use this dataset to evaluate the performance of hand pose estimation under challenging scenarios.

~

\noindent\textbf{MSCOCO~\cite{mscoco}:}\quad
It is a large-scale in-the-wild 2D human pose dataset consisting of over 150k instances. We incorporate its train set for training.

\section{Ablation Experiments.}
\label{sec:supp_ablation}

\begin{table}[h]
   \begin{center}
   \caption{\textbf{Naive \textit{vs.} Adaptive} with different input joints on body-only mesh recovery. MPJPE of 24 joints is reported. Adaptive HybrIK is more robust to the noise.}
   \label{tab:ablation_noise}
   \resizebox{\linewidth}{!}
   {%
       \begin{tabular}{l|c|c|c|c}
           \toprule
           ~& GT Joints & $\pm 10$ mm & $\pm 20$ mm & $\pm 30$ mm \\
           \midrule
           Naive HybrIK & 0.1 & 43.8 & 74.1 & 100.2 \\
           Adaptive HybrIK & 0.1 & 21.1 & 43.6 & 67.5 \\
           \bottomrule
       \end{tabular}
   }
   \end{center}
\end{table}

\noindent\textbf{Robustness of HybrIK to Noisy Joint Positions.} In the main paper, we evaluate the robustness of HybrIK in whole-body scenarios. Here, we further report the evaluation of robustness on body-only scenarios. We use the same evaluation protocol as in the main paper. We randomly add Gaussian noise to the 3D joint positions with different standard deviations. The results are shown in Tab.~\ref{tab:ablation_noise}. We can see that adaptive HybrIK is more robust to the noise than naive HybrIK.

\begin{table}[h]
   \begin{center}
   \caption{\textbf{Reconstruction error with different shape parameters $\beta$} on the AGORA validation set.}
   \label{tab:beta}
   {%
         \begin{tabular}{l|cc|cc|cc}
         \toprule
         & \multicolumn{2}{c}{GT $\beta$} & \multicolumn{2}{c}{Estimated $\beta$} & \multicolumn{2}{c}{Zero $\beta$} \\
         \cmidrule(lr){2-3} \cmidrule(lr){4-5} \cmidrule(lr){6-7}
         ~& MPJPE & PVE & MPJPE & PVE & MPJPE & PVE  \\
         \midrule
         Error & 65.4 & 74.9 & 67.9 & 77.1 & 73.4 & 83.4 \\
         \bottomrule
         \end{tabular}
   }
   \end{center}
\end{table}

\begin{table*}[!t]
   \begin{center}
   \caption{\textbf{Error correction capability of HybrIK} on the Human3.6M, 3DPW, and AGORA datasets.}
   \label{tab:correction}
   {%
         \begin{tabular}{l|cc|cc|cc}
         \toprule
         & \multicolumn{2}{c}{Human3.6M} & \multicolumn{2}{c}{3DPW} & \multicolumn{2}{c}{AGORA}\\
         \cmidrule(lr){2-3} \cmidrule(lr){4-5} \cmidrule(lr){6-7}
         ~& Predicted Pose & HybrIK & Predicted Pose & HybrIK & Predicted Pose & HybrIK \\
         \midrule
         MPJPE (24 \textit{jts}) $\downarrow$ & 50.9 & 48.1 & 78.9 & 74.8 & 73.1 & 66.3 \\
         \bottomrule
         \end{tabular}
   }
   \end{center}
\end{table*}

~

\noindent\textbf{Effect of $\beta$.}
In this experiment, we examine the impact of shape parameters $\beta$ on the AGORA validation set. As shown in Tab.~\ref{tab:beta}, using the ground-truth $\beta$ yields a $2$ mm improvement in MPJPE and PVE, while using zero $\beta$ results in a $6$ mm error. It shows that our model also gives an accurate body shape estimation.

~


\noindent\textbf{Error correction capability of HybrIK.}
In this experiment, we investigate the error correction capability of body-only HybrIK on the 3DPW~\cite{3dpw}, Human3.6M~\cite{h36m}, and AGORA~\cite{patel2021agora} datasets. Quantitative results are reported in Tab.~\ref{tab:correction}. They demonstrate that HybrIK can leverage the structural information embedded in the statistical body model to rectify unrealistic body joints derived from off-the-shelf 3D keypoint estimation approaches. The error correction capability of HybrIK is more pronounced on the AGORA dataset, which poses significant challenges for pose and joint estimation.

\section{Qualitative Results}
\label{sec:qualitative}
\begin{figure*}[ht]
    \begin{center}
        \includegraphics[width=\linewidth]{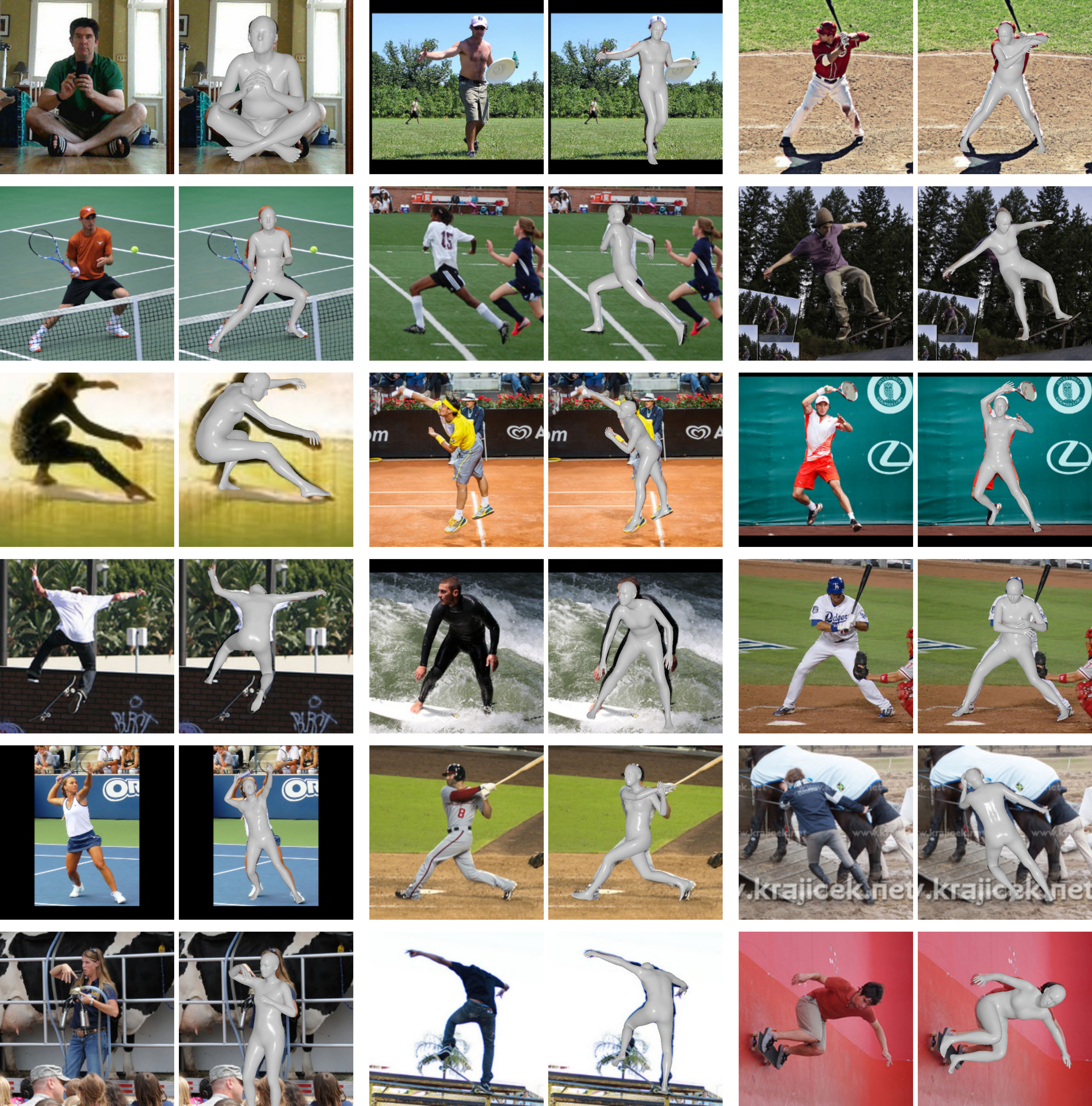}
    \end{center}
    \caption{\textbf{Qualitative results of body-only mesh recovery on challenging poses.}
    }
    \label{fig:qualitative_body_supp}
\end{figure*}

\begin{figure*}[ht]
    \begin{center}
        \includegraphics[width=\linewidth]{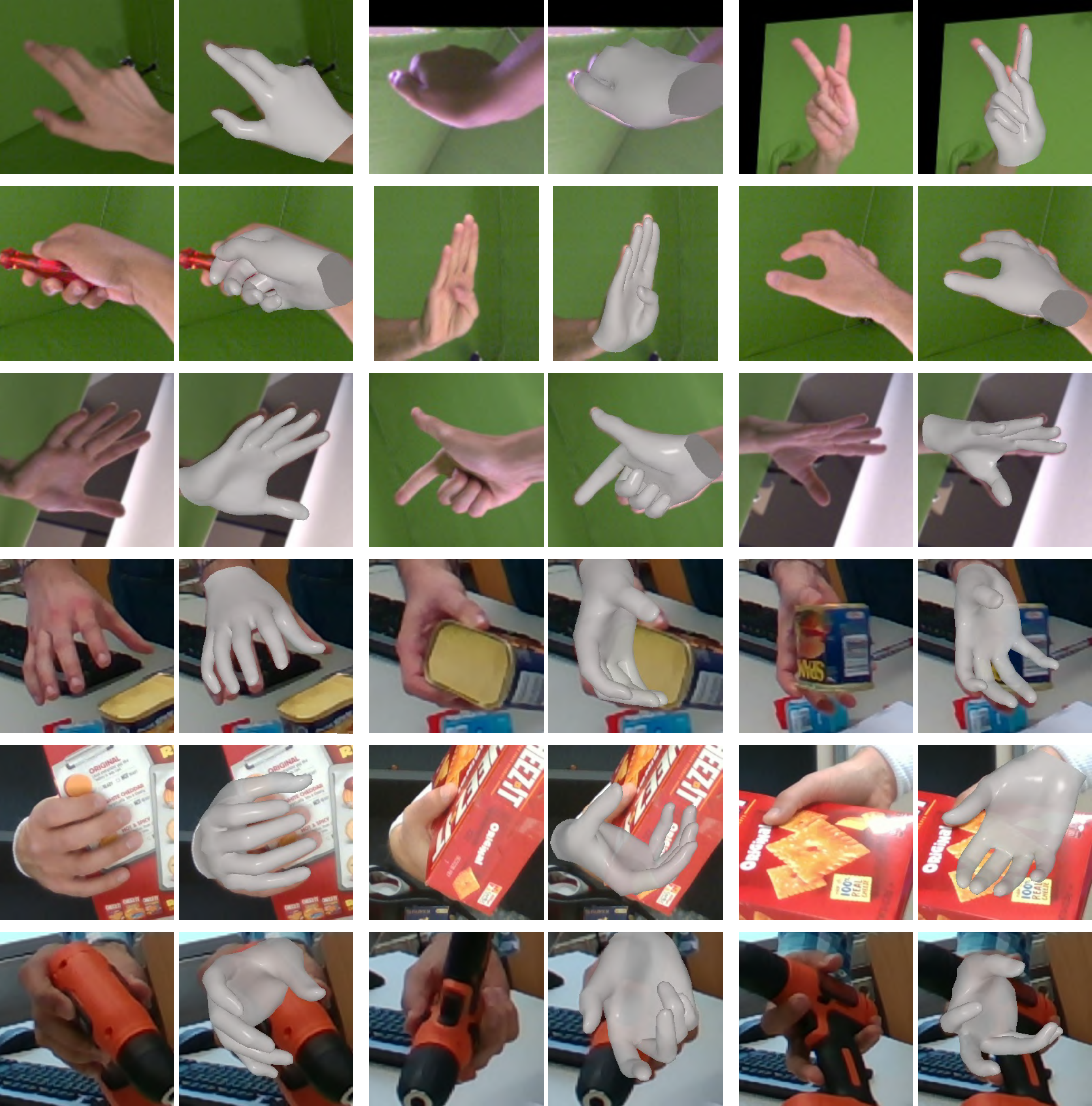}
    \end{center}
    \caption{\textbf{Qualitative results of hand-only mesh recovery on the FreiHAND dataset (rows 1-3) and the HO3D-v2 dataset (rows 4-6).}
    }
    \label{fig:qualitative_hand}
\end{figure*}

\begin{figure*}[ht]
    \begin{center}
        \includegraphics[width=\linewidth]{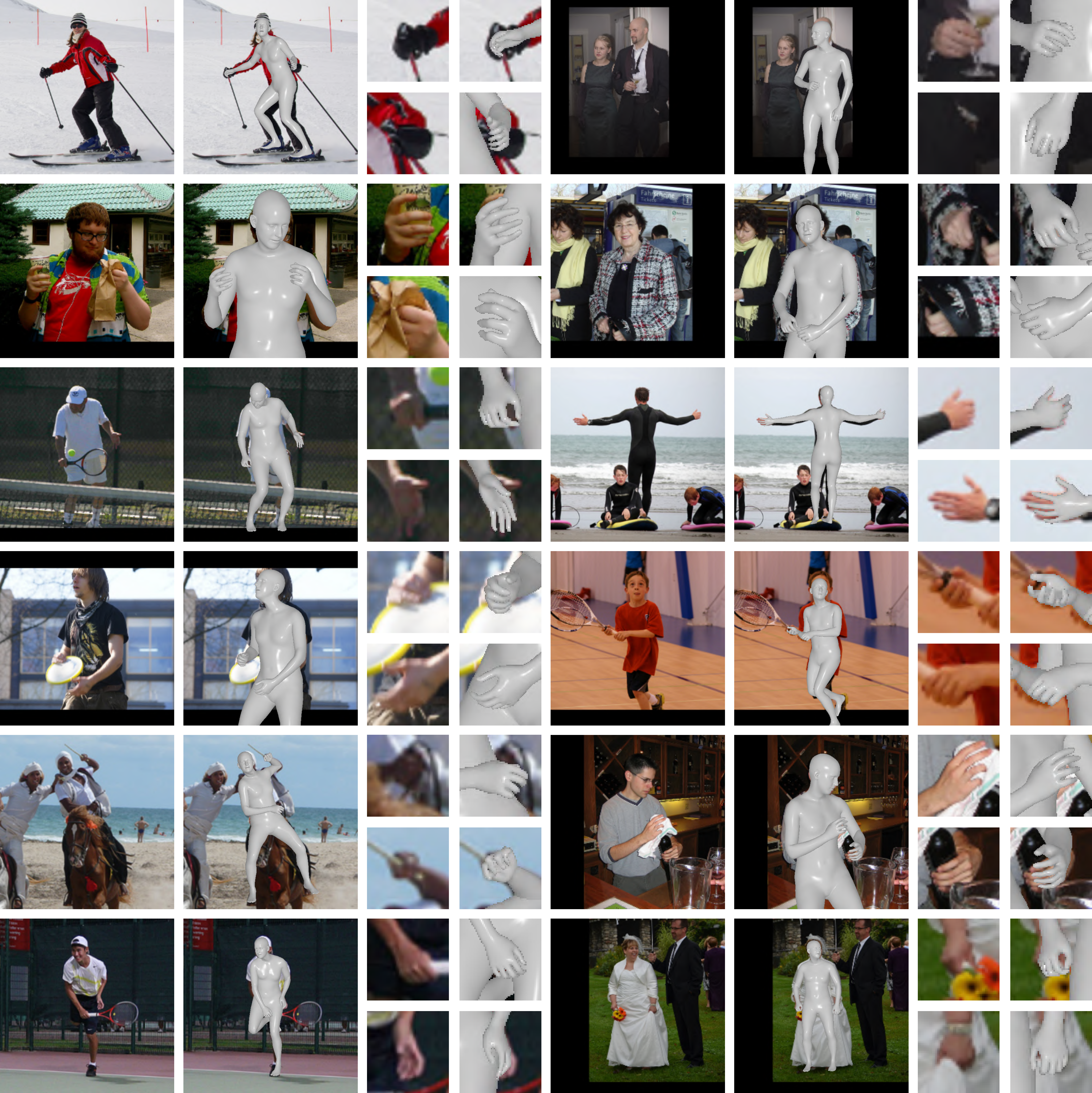}
    \end{center}
    \caption{\textbf{Qualitative results of whole-body mesh recovery on challenging poses.}
    }
    \label{fig:qualitative_wholebody}
\end{figure*}

\begin{figure*}[ht]
    \begin{center}
        \includegraphics[width=\linewidth]{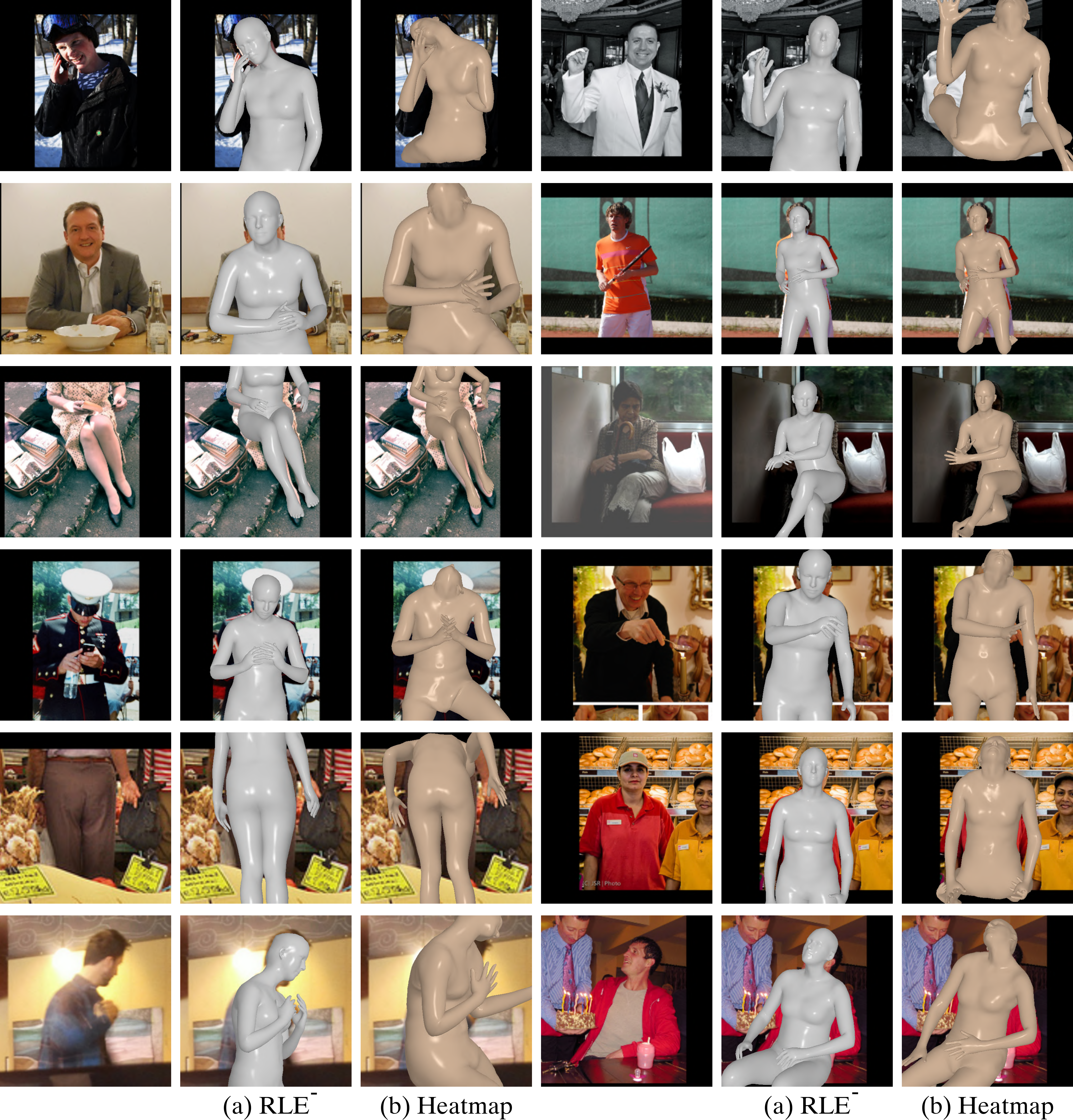}
    \end{center}
    \caption{\textbf{Qualitative comparisons between RLE$^-$-based and heatmap-based body-only mesh recovery.}
    }
    \label{fig:qualitative_truncate}
\end{figure*}

\begin{figure*}[ht]
    \begin{center}
        \includegraphics[width=\linewidth]{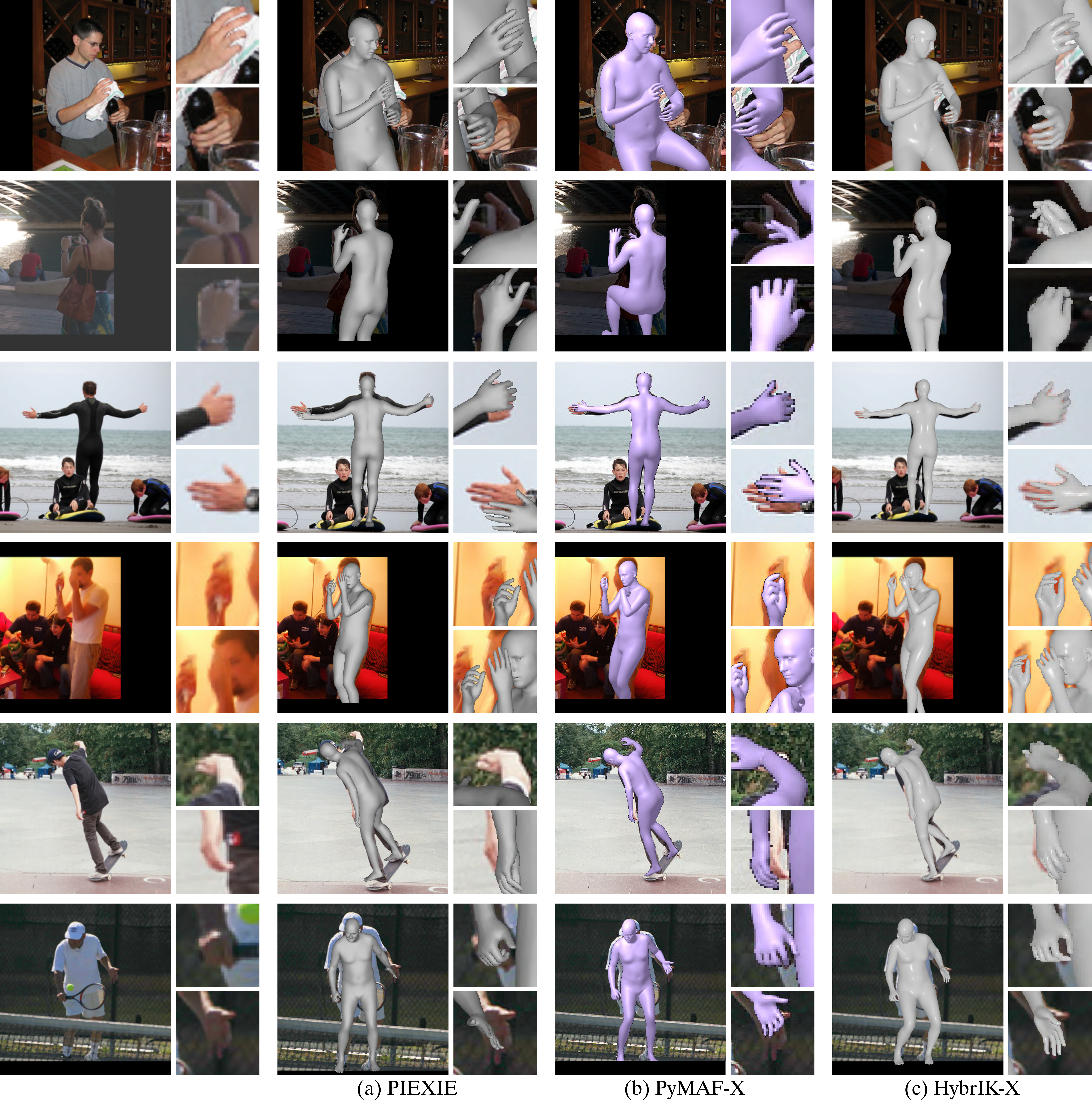}
    \end{center}
    \caption{\textbf{Qualitative comparisons with state-of-the-art approaches.}
    }
    \label{fig:qualitative_compare}
\end{figure*}

Additional qualitative results for body-only, hand-only, and whole-body scenarios are presented in Fig.~\ref{fig:qualitative_body_supp}, \ref{fig:qualitative_hand}, and \ref{fig:qualitative_wholebody}, respectively. Qualitative comparisons between heatmap-based backbone and our regression-based backbone are displayed in Fig.~\ref{fig:qualitative_truncate}. Qualitative comparisons with state-of-the-art approaches are presented in Fig.~\ref{fig:qualitative_compare}. More results can be found in our project page \href{https://jeffli.site/HybrIK-X/}{https://jeffli.site/HybrIK-X/}.

\end{appendices}

\clearpage

\bibliographystyle{IEEEtran}
\bibliography{egbib}

%



\end{document}